\pdfoutput=1

\documentclass[11pt,table,dvipsnames]{article}

\usepackage{acl}

\usepackage{times}
\usepackage{latexsym}

\usepackage[T1]{fontenc}

\usepackage[utf8]{inputenc}

\usepackage{microtype}

\usepackage{inconsolata}

\usepackage{graphicx}
\usepackage{booktabs} 
\usepackage{multirow, makecell, caption}
\usepackage{arydshln} 
\usepackage{xspace}
\usepackage{enumitem}
\usepackage{tabularx}
\usepackage{float}
\usepackage{pifont}
\usepackage{amsmath,amssymb,amsfonts}
\usepackage[linesnumbered,ruled,vlined]{algorithm2e} 
\usepackage{algorithmicx}

\SetCommentSty{mycommfont}

\makeatletter
\def\adl@drawiv#1#2#3{%
        \hskip.5\tabcolsep
        \xleaders#3{#2.5\@tempdimb #1{1}#2.5\@tempdimb}%
                #2\z@ plus1fil minus1fil\relax
        \hskip.5\tabcolsep}
\newcommand{\cdashlinelr}[1]{%
  \noalign{\vskip\aboverulesep
           \global\let\@dashdrawstore\adl@draw
           \global\let\adl@draw\adl@drawiv}
  \cdashline{#1}
  \noalign{\global\let\adl@draw\@dashdrawstore
           \vskip\belowrulesep}}
\makeatother

\title{Laying the Foundation First? Investigating the Generalization \\ from Atomic Skills to Complex Reasoning Tasks}

\author{Yuncheng Huang\textsuperscript{\rm $\spadesuit$}, 
 Qianyu He\textsuperscript{\rm $\spadesuit$}, 
 Yipei Xu\textsuperscript{\rm $\spadesuit$}, 
 Jiaqing Liang\textsuperscript{\rm $\heartsuit$}\thanks{~~Corresponding author.}, 
 Yanghua Xiao\textsuperscript{\rm $\spadesuit\diamondsuit$}\footnotemark[1]\\
\textsuperscript{\rm $\spadesuit$}Shanghai Key Laboratory of Data Science, School of Computer Science, Fudan University\\
\textsuperscript{\rm $\heartsuit$}School of Data Science, Fudan University\\
\textsuperscript{\rm $\diamondsuit$}Fudan-Aishu Cognitive Intelligence Joint Research Center\\
\texttt{\{yunchenghuang22, qyhe21, ypxu22\}@m.fudan.edu.cn}\\
\texttt{\{liangjiaqing, shawyh\}@fudan.edu.cn}
}

\begin{document}
\maketitle
\begin{abstract}
Current language models have demonstrated their capability to develop basic reasoning, but struggle in more complicated reasoning tasks that require a combination of atomic skills, such as math word problem requiring skills like arithmetic and unit conversion.
Previous methods either do not improve the inherent atomic skills of models or not attempt to generalize the atomic skills to complex reasoning tasks.
In this paper, we first propose a probing framework to investigate whether the atomic skill can spontaneously generalize to complex reasoning tasks.
Then, we introduce a hierarchical curriculum learning training strategy to achieve better skill generalization.
In our experiments, we find that atomic skills can not spontaneously generalize to compositional tasks.
By leveraging hierarchical curriculum learning, we successfully induce generalization, significantly improve the performance of open-source LMs on complex reasoning tasks.
Promisingly, the skill generalization exhibit effective in cross-dataset and cross-domain scenarios.
Complex reasoning can also help enhance atomic skills. 
Our findings offer valuable guidance for designing better training strategies for complex reasoning tasks.

\end{abstract}
\section{Introduction}

\begin{figure}
    \centering
    \includegraphics[width=0.5\textwidth, height=0.35\textwidth]{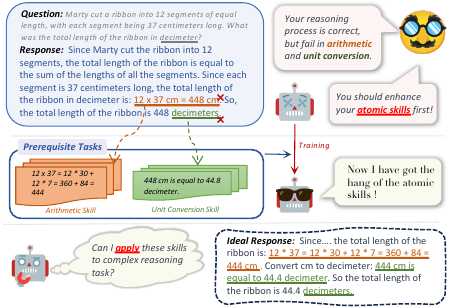}
    \caption{An example of LMs' deficiencies on atomic skills when solving complex reasoning tasks. While these atomic skills can be improved through skill training, it remains uncertain whether language models can apply enhanced skills to complex tasks.}
    \label{fig:motivation}
\end{figure}

Current language models (LMs) have demonstrated their capability in a variety of reasoning tasks~\citep{huang-chang-2023-towards,wei2022chain}.
However, they struggle in more complex tasks that require the combination of various atomic skills, such as solving math word problem (MWP, ~\citealp{cobbe2021gsm8k,patel-etal-2021-nlp}) requiring arithmetic~\citep{liu2023goat,nogueira2021investigating,muffo-etal-2022-evaluating} and unit conversion~\citep{park-etal-2022-language} skills.
Previous study argue that the inferior performance of current LMs in solving complex reasoning tasks is primarily attributed to their deficiency in atomic skills.
As shown in Fig.~\ref{fig:motivation}~(top), despite following a correct reasoning process, the models still yield incorrect solutions due to errors in arithmetic and unit conversion skills.

Recent studies attempt to address this issue through skill enhancement, but there are still limitations.
Some approach involves introducing external tools~\citep{imani-etal-2023-mathprompter,schick2023toolformer}, validators~\citep{grace2023} or knowledge bases~\citep{lewis2020retrieval} to assist atomic skills.
These methods rely on external support but do not inherently improve the atomic skills of the model itself.
Other studies promote the performance through multitask learning~\citep{Chen2023SkillitAD,kim-etal-2023-taskweb}. 
They argue that skill improvement can be implicitly achieved through the transfer effect between tasks.
However, they neither specify which skills are improved nor quantitatively assess the performance gains from skill improvements.
The skill enhancement is implicit and unobservable, and the relationships between tasks are inexplainable.
The most related studies individually improve particular skills by integrating specific knowledge~\cite{park-etal-2022-language} or by fine-tuning with specialized crafted Chain-of-Thought~\cite{liu2023goat}.
However, these studies tend to train a specialized models that proficient in atomic skills rather than enhance atomic skills while maintaining the original capabilities of the model.
Moreover, they do not investigate whether the enhancement of skills can be generalized to complex tasks.

We argue that skill enhancement can \textit{generalize} to complex tasks, as the response format for complex tasks is a \textit{composition} of atomic skills.
For instance, in Fig.~\ref{fig:motivation}, the response to the MWP compose arithmetic and unit conversion skills, which are respectively corresponding to the text segments ``12$\times$37=448 cm'' and ``448 decimeters''.
The precision of complex reasoning tasks is significantly influenced by the mastery of skills.
In this case, if both skills are improved, the response would turn out to be correct.
Language models have been proved to individually improve their skills through specialized training~(Fig.~\ref{fig:motivation}, middle).
What we are particularly interested in is whether models can \textit{apply} the enhanced skills to complex tasks~(Fig.~\ref{fig:motivation}, bottom), referred to as \textbf{skill generalization} in this paper.
It is crucial to highlight that our research objective is fundamentally different from multitasking as we explicitly define skills.
Furthermore, due to the \textit{composability} between skills and complex tasks, this generalization effect should be observable and explainable.

In this work, we investigate the mechanism of skill generalization through empirical experimentation on MWP.
We aim to answer two key questions:  \textit{Can atomic skills spontaneously generalize to complex reasoning tasks?} \textit{How can we maximize the skill generalization effectiveness?}
First, we propose a probing framework to investigate the skill generalization mechanism on complex reasoning tasks.
We select two essential atomic skills in MWP for probing: arithmetic and unit conversion.
Then, we specifically design prerequisite tasks to enhance atomic skills and construct corresponding datasets through automated methods.
Moreover, inspired by hierarchical curriculum design in pedagogy~\citep{white1974past,2008STUDENT}, we propose a two-stage training strategy named hierarchical curriculum learning to maximize skill generalization.
The first stage is skill training, which involves continuous learning on prerequisite tasks, enabling LMs to enhance atomic skills while maintaining their original problem-solving abilities.
The second stage is applied learning, where language models learn to apply skills to complex reasoning tasks.
Finally, we carry out experiments across different models and perform detailed analyses.

In our experiments, we observe that (1) atomic skill can not spontaneously generalize to complex reasoning tasks, but can be induced to generalize through hierarchical curriculum learning.
(2) A strong foundation laid in skill learning is crucial for effectiveness of LMs on complex reasoning tasks.
(3) Skill enhancement exhibits a cross-dataset and cross-domain generalization effect. 
(4) Conversely, complex reasoning task can also help enhance the atomic skills.
We attribute this to the \textit{composability} between skills and complex tasks.

Our contributions can be summarized as follows:
\begin{itemize}[itemsep=0.8pt]
    \item To our best knowledge, we are the first to investigate the generalization from atomic skills to complex reasoning tasks.
    \item We propose a probing framework to investigate the spontaneity and effectiveness of skill generalization.
    \item We propose a hierarchical curriculum learning training strategy to induce skill generalization. Our experiments demonstrate the effectiveness of this strategy in achieving better skill generalization.
\end{itemize}

\section{Related Work}

\paragraph{Task Generalization}
Cross-task generalization refers to effectively apply previously learned knowledge and skills from source task to new target tasks~\citep{talmor-berant-2019-multiqa,khashabi-etal-2020-unifiedqa,ye-etal-2021-crossfit}.
Recent studies attain significant success in cross-task generalization by employing a multi-tasking approach~\citep{sanh2021multitask,wei2021finetuned,kim-etal-2023-taskweb}.
\citet{Chen2023SkillitAD} argues that the effectiveness of generalization stems from the implicit  skill transfer between tasks and seeks to find an optimal sequence to maximize the effect.
Our research differs from the aforementioned studies in that we explicitly predefine source and target tasks that possess \textit{composability} in format.
Moreover, our research does not depend on massive tasks but emphasizes generalization from atomic skills to complex reasoning tasks.

\paragraph{Compositional Generalization}
Compositional generalization research primarily focus in semantic parsing~\cite{lake2018generalization,keysers2019measuring,kim-linzen-2020-cogs}. 
They explore generalizing simple data to complex data through composition within an inter-dataset distribution.
In contrast, our study explores cross-dataset generalization, especially skill generalization in complex reasoning tasks.

\paragraph{Atomic Skill Learning}
Numerous studies focus on individually enhancing specific skills.
~\citet{liu2023goat} enhance arithmetic skill of LMs by specialized designed COT prompting.
~\citet{huang2023enhancing} improve unit conversion skills through dimensional perception pretraining tasks.
However, these studies do not investigate generalizing the enhanced skills to complex reasoning tasks.

\paragraph{Curriculum Learning}
Curriculum learning suggests that a structured and progressively challenging learning path can improve the learning effectiveness~\cite{bengio2009curriculum,wu2020curricula}.
Previous work focus on ordinal training on a single task based on the difficulty of the data~\cite{jiang2015self,xu-etal-2020-curriculum,elgaar-amiri-2023-hucurl}.
Our research advances the field by applying hierarchical curriculum learning to multitasks guided by \textit{composability} among these tasks and investigates their generalization effects.

\section{Method}

In this section, we first propose a probing framework to investigate generalization from atomic skills to complex reasoning tasks (\S~\ref{sec:probe}).
Then, we propose a hierarchical curriculum learning strategy to maximize the generalization effect (\S~\ref{sec:hcl}). The framework is shown in Fig.~\ref{fig:framework}.

\subsection{Skill Generalization Probing}
\label{sec:probe}

\subsubsection{Task Selection}

We chose math word problem (MWP, ~\citealp{cobbe2021gsm8k,patel-etal-2021-nlp}) as the investigated task, as it is a common-used benchmark for complex reasoning and the correctness can be objectively assessed.
We select arithmetic and unit conversion as atomic skills because LMs display weaknesses in these skills when addressing MWP~\citep{imani-etal-2023-mathprompter,schick2023toolformer,huang2023enhancing}.
To obtain a model proficient in skills, we need to design prerequisite tasks and conduct skill training first (\S~\ref{sec:skill}).
After that, we can investigate the skill generalization on the enhanced model (\S~\ref{sec:how})

\begin{figure}[t]
    \centering
    \includegraphics[width=0.44\textwidth, height=40mm]{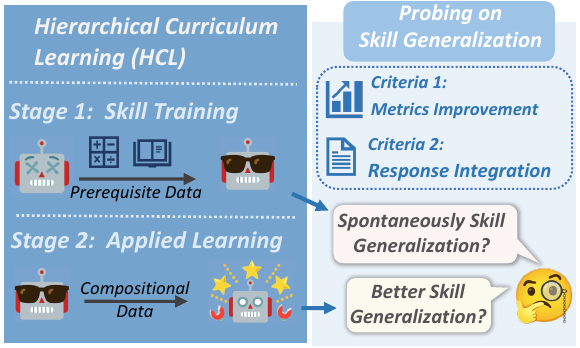}
    \caption{Framework of our method. The right part is our probing approach. The left part describes the model training stages in hierarchical curriculum learning.}
    \label{fig:framework}
\end{figure}

\definecolor{darkorange}{RGB}{204, 102, 0}
\definecolor{darkgreen}{RGB}{0, 100, 0}

\begin{table*}[t]
    \small
    \centering
        \begin{tabular}{cccl}
        \toprule
\multicolumn{2}{c}{\textbf{Task}} & \textbf{Type}& \multicolumn{1}{c}{\textbf{Example}} \\
\midrule
\multirow{7}{*}{\rotatebox{90}{\textbf{Prerequisite Tasks}}} & \multirow{4}{*}{Arithmetic} & \texttt{AddSub}  & \textit{5520.8 + 1.34 = 5522.14;} \ \  \textit{5494 + 26.8 + 1.34 = 5520.8 + 1.34 = 5522.14;}\\
& & \texttt{S-Mul} & \textit{12 * 40 = 480;} \ \ \textit{12 * 3 = 36;} \ \ \textit{12 * 0.5 = 6;}  \ \ \textit{12 * 0.01 = 0.12} \\

&  & \multirow{1}{*}{\texttt{C-Mul}} & \multirow{1}{*}{\parbox{11cm}{\textcolor{darkorange}{\textbf{\textit{12 * 43.5 = 12 * 40 + 12 * 3 + 12 * 0.5 = 480 + 36 + 6 = 516 + 6 = 522}}}}} \\
& & \texttt{S-Div} & \textit{123 / 2 = 61.5;} \ \ \textit{214 / 3 = 80.33} \ \ \textit{123 / 10 = 12.3} \\
 \cmidrule{2-4}
& \multirow{3}{*}{Unit Conversion} & Length  & \textcolor{darkgreen}{\textbf{\textit{522 meter is equal to 0.522 kiloteters}}}. \ \ \ \textit{Two inches is equal to 5.08 centimeters.}\\
&  & Time &  \textit{1 hour is equal to 60 minutes}. \ \ \ \textit{3 hours is equal to 10800 seconds}.\\
& & Speed & \textit{1 m/s is equal to 3.6 km/h.} \ \ \ \textit{72 kilometer per hour is equal to 20 meter per second.}\\
 \midrule
\multirow{8}{*}{\rotatebox{90}{\textbf{Complex Tasks}}}  & \multirow{3}{*}{Math Word Problem} & \multirow{3}{*}{-} & \multirow{3}{*}{\parbox{11cm}{\textbf{Question}: James decides to run 3 sprints 4 times a week. He runs 43 meters each sprint. How many total \underline{meters} does he run a week?\\
\textbf{Response}: He sprints 3 * 4 = 12 times. So he runs 12 * 43 = 516 meters a week.}} \\
&  & & \\
& & & \\
 \cmidrule{2-4}
& \multirow{5}{*}{Applied Learning} & \multirow{5}{*}{Mixture} & \multirow{4}{*}{\parbox{11cm}{\textbf{Question}: James decides to run 3 sprints 4 times a week. He runs 43.5 meters each sprint. How many total \underline{kilometers} does he run a week? \\
\textbf{Response}:He sprints 3 * 4 = 12 times. So he runs \textcolor{darkorange}{\textbf{\textit{12 * 43.5 = 12 * 40 + 12 * 3 + 12 * 0.5 = 480 + 36 + 6 = 516 + 6 = 522}}} meters a week. \textcolor{darkgreen}{\textbf{\textit{522 meters is equal to 0.522 kilometers}}}. So the answer is 0.522.}} \\
& & & \\
& & & \\
& & & \\
& & & \\
 
        \bottomrule
        \end{tabular}
    \caption{Examples for prerequisite tasks and complex reasoning tasks. The response for compositional tasks presents arithmetic and unit conversion skill and they are highlighted in orange and green respectively. \texttt{S-} refers to simple operation where the significant digit of the second number is 1. \texttt{C-} refers to complex operation.
    }
    \label{tab:data_example}
\end{table*}

\subsubsection{Probing Skills}
\label{sec:skill}

\paragraph{Arithmetic Skill.}
Arithmetic skill refer to perform operations among numbers such as addition, subtract, multiplication and division.
Most current LMs suffer from inaccurate arithmetic due to lacking specialized skill-oriented training~\citep{liu2023goat}.
We design a prerequisite task for arithmetic and construct the corresponding dataset.
By training on prerequisite tasks, we can enhance the arithmetic skills.
The arithmetic data encompasses a variety of difficulties, including different operation hops, operation types, value types and significant digits.
For simple operations, we require the model to directly provide the arithmetic result.
For complex operations, we design Chain-of-Thought responses, following~\citet{liu2023goat}, due to the challenges in directly deriving the answers for these tasks.
As shown in the example in Tab.~\ref{tab:data_example}, when answering ``12 * 43.5'', we require the model to present the process of splitting, expansion, producting, and adding term by term before providing the final answer.

\paragraph{Unit Conversion Skill.}
Similar to arithmetic, unit conversion is necessary when dealing with values of different units in MWP.
Current LMs lack sufficient knowledge of units, making it difficult to accurately perform unit conversions~\citep{huang2023enhancing}.
Therefore, we also propose a prerequisite task and corresponding training data for unit conversion.
We first extract all quantity types in MWP based on a comprehensive unit knowledge base DimUnitKB~\citep{huang2023enhancing}.
As shown in Tab.~\ref{tab:data_example}, the units involved in MWP include seven quantity types such as \texttt{length}, \texttt{time}, \texttt{speed}, etc. 
Then we construct the unit conversion dataset for unit pair under the same quantity type.
For example, ``meters'' and ``centimeters'' are both denote length, so it can be naturally stated as ``\textit{1 meter is equal to 100 centimeters}".
We detail the constrution method in Appendix~\ref{sec:ap_unit_data}.

\subsubsection{Skill Training (ST)}
\label{sec:skill-train}

Since the data for arithmetic and unit conversion are both automatically constructed, we can generate them in large quantities.
It is straight-forward to enhance the atomic skills into language models through continuous training.
However, continuous training may lead to catastrophic forgetting~\citep{mccloskey1989catastrophic}.
To address this, we employ the \textit{replay strategy}~\citep{ke2022continual} that is widely used in continuous training.
We retain some training examples from MWP and mix them\footnote{We discuss the mixing ratio in Appendix~\ref{sec:ratio}.} with prerequisite task data to ensure the model retains its original problem-solving abilities in skill training.
We conduct individual training for each skills as well as training with a mixture of skills.

\subsubsection{\textit{How to determine whether skill generalization has been achieved?}}
\label{sec:how}
Skill generalization refers to being able to \textit{apply} skills learned from prerequisite tasks to complex reasoning tasks.
Therefore, we can assess this by testing the skill-enhanced model in \S~\ref{sec:skill} on MWP.
We can consider the following aspects.

\noindent \textbf{Metrics Improvement:} Skills improvement can fix mistakes caused by the deficiency of atomic skills of language models when solving reasoning tasks.
Therefore, ideally, skill generalization should be reflected in an improvement in metrics.
\noindent \textbf{Response Integration:} 
As seen in Tab.~\ref{tab:data_example}, the format we use for atomic skills in prerequisite tasks differs from how the original model performs these skills.
Therefore, we can assess by determining whether there has been an implementation of response integration.
For example, successful skill generalization should involve performing \texttt{C-Mul} in a Chain of Thought (COT) format rather than providing the answer directly.
\subsection{Hierarchical Curriculum Learning (HCL)}
\label{sec:hcl}

Probing experiments show that atomic skills can not spontaneously generalize to complex tasks (results are detailed in \S~\ref{sec:rq1}).
Therefore, we propose hierarchical curriculum learning (HCL) to induce skill generalization.

Our approach is primarily inspired by hierarchical curriculum design in pedagogy~\citep{white1974past,2008STUDENT}.
In most of education system, student complete prerequisite course before enrolling in a more advanced course~\citep{huang2005prerequisite}.
Prerequisites are to ensure that students possess necessary foundational knowledge and skills and advanced courses enable students to learn how to apply these skills in complex scenarios~\citep{rovick1999accurate}.
Likewise, we design a two-stage hierarchical curriculum learning framework in our setting, shown in Fig.~\ref{fig:framework} left.
The first stage is skill training, which has already been implemented (\S~\ref{sec:skill-train}). 
We introduce the second phase of \textit{applied learning} to enable LMs to apply their acquired skills to complex tasks.

\subsubsection{Applied Learning (AL)}

In this stage, we first construct compositional data (shown in Tab.~\ref{tab:data_example} bottom) for applied learning.
Next, we further train the model in the first stage with compositional data.
In Tab.~\ref{tab:data_example}, the response of the original MWP directly provides the result when performing arithmetic.
Moreover, the responses usually do not show the process of converting units.
We incorporates the response format from prerequisite tasks into the problem-solving process for MWPs, aiming to induce the model to apply atomic skills to complex tasks. 
We detail the data construction method in Appendix~\ref{sec:ap_composition}.

\section{Experimental Settings}

\subsection{Evaluation Datasets.}
We choose GSM8K~\cite{cobbe2021gsm8k}, a widely used benchmark for complex multi-step reasoning, requiring arithmetic and unit conversion skills.
We compile statistics of arithmetic and unit conversion in the test set and observe that it lacks comprehensiveness in terms of difficulty and knowledge coverage.
Therefore, we enhance the difficulty of GSM8K test set to demonstrate skill generalization more significantly. We denote the origin dataset as \texttt{RAW} and the augmented dataset as \texttt{HARD}.

\begin{figure}
    \centering
    \includegraphics[width=0.5\textwidth, height=0.16\textwidth]{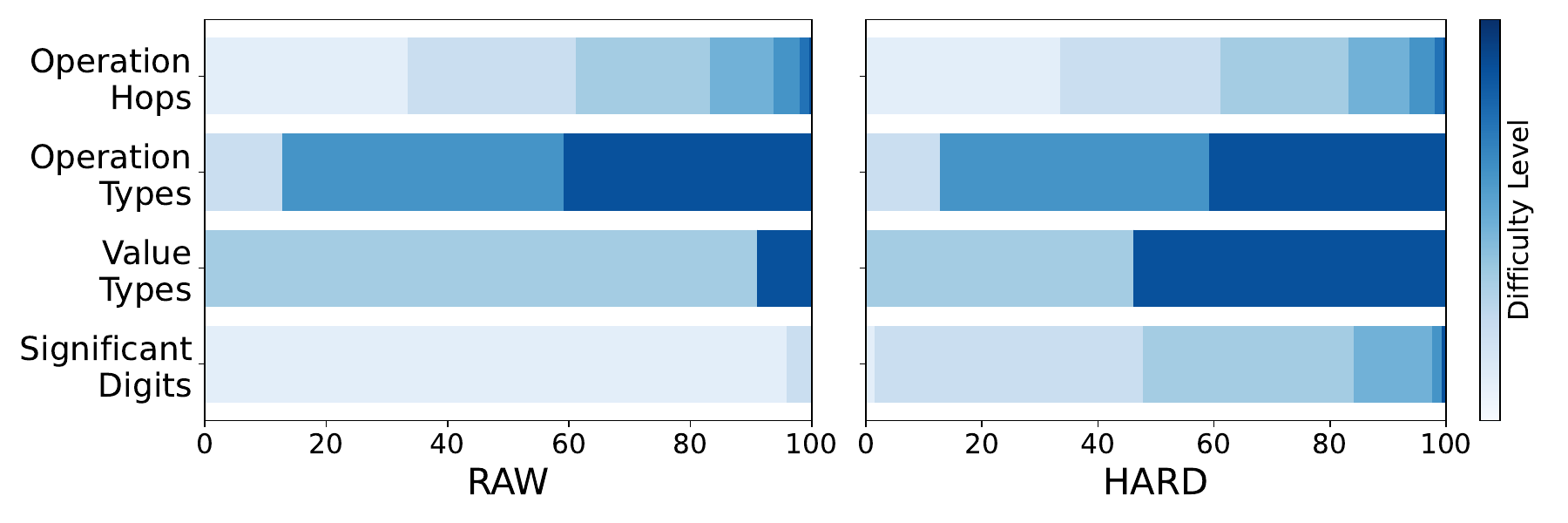}
    \caption{The distribution of data difficulty across four dimensions. Darker colors mean greater difficulty and larger areas mean more data.}
    \label{fig:arith_statistic}
\end{figure}

\paragraph{Arithmetic Augmentation.} 
The difficulty of arithmetic skills can be assessed from four dimensions: operation hops, operation type, value type, and significant digit.
\texttt{RAW} has reasonable settings in the first three dimensions, but its inclusion of short significant digit resulting in lower demands on arithmetic skills.
Therefore, we extend the significant digits in \texttt{RAW} without changing the logic of the original problems.
In Tab.~\ref{tab:statistics-arithaug-metric} we showcase the performance of LLaMA-2~\citep{touvron2023llama} on the test set before and after enhancement. 
The difficulty of these three operations increases progressively, but testing on \texttt{RAW} can not distinguish the difficulty.
The enhanced test set aligns with this difficulty gradient, demonstrating the effectiveness of the enhanced dataset.

\begin{table}[t]
    \centering
    \small
    \begin{tabular}{lcccc}
    \toprule
        \multicolumn{1}{l}{\textbf{Dataset}} & \textbf{Operartion} & \texttt{AddSub} & \texttt{Mixmul} & \texttt{MixAll} \\
    \midrule
        \multirow{2}{*}{GSM8K$_{RAW}$}
               & 2-Hop
& 43.10 & 44.24 & 34.88 \\
             & 3-Hop
& 22.72 & 35.00 & 14.32 \\
        \midrule
        \multirow{2}{*}{GSM8K$_{HARD}$}
               & 2-Hop
& 31.90 & 15.20 & 5.30 \\
             & 3-Hop
& 21.50 & 8.04 & 2.63 \\
    \bottomrule
    \end{tabular}
    \caption{Arithmetic accuracy(\%) of LLaMA-2 on \texttt{RAW} and \texttt{HARD}. \texttt{AddSub} involves only addition and subtraction operations, \texttt{MixMul} includes multiplication, and \texttt{MixAll} involves all operations.}
    \label{tab:statistics-arithaug-metric}
\end{table}

\newcolumntype{a}{>{\columncolor{BlueGreen!10}}c}
\newcolumntype{b}{>{\columncolor{Green!10}}c}
\newcolumntype{d}{>{\columncolor{Gray!10}}c}
\newcolumntype{q}{>{\columncolor{Blue!10}}c}

\begin{table*}[t]
    \small
    \centering
        \begin{tabular}{ccaaaabbbbqqqq}
        \toprule
        \multirow{3}{*}{\textbf{Model}} & \multirow{3}{*}{\textbf{Method}} & \multicolumn{4}{c}{\textbf{Arithmetic}} & \multicolumn{4}{c}{\textbf{Unit Conversion}} & \multicolumn{4}{c}{\textbf{Mixture}} \\
        \cmidrule(lr){3-6}  \cmidrule(lr){7-10} \cmidrule(lr){11-14}
        & & \multicolumn{2}{c}{\textbf{Zero-shot}} &  \multicolumn{2}{c}{\textbf{Few-shot}} & \multicolumn{2}{c}{\textbf{Zero-shot}} &  \multicolumn{2}{c}{\textbf{Few-shot}} & \multicolumn{2}{c}{\textbf{Zero-shot}} &  \multicolumn{2}{c}{\textbf{Few-shot}} \\
        \cmidrule(lr){3-4} \cmidrule(lr){5-6} \cmidrule(lr){7-8} \cmidrule(lr){9-10} \cmidrule(lr){11-12} \cmidrule(lr){13-14}
         & & \multicolumn{1}{c}{\texttt{RAW}}  &  \multicolumn{1}{c}{\texttt{HARD}}  
         & \multicolumn{1}{c}{\texttt{RAW}} & \multicolumn{1}{c}{\texttt{HARD}}
         & \multicolumn{1}{c}{\texttt{RAW}} & \multicolumn{1}{c}{\texttt{HARD}}
         & \multicolumn{1}{c}{\texttt{RAW}} &\multicolumn{1}{c}{\texttt{HARD}}
         & \multicolumn{1}{c}{\texttt{ARITH}} & \multicolumn{1}{c}{\texttt{UNIT}}
         & \multicolumn{1}{c}{\texttt{ARITH}} & \multicolumn{1}{c}{\texttt{UNIT}} \\
\midrule

\multirow{5}{*}{LLaMa-2-7B} & Vanilla 
& 22.97  & 8.67 & 23.73 & 10.03 & 23.12 & 1.72 & 23.50 & 6.90 & 8.67 & 1.72 & 10.03 & 6.90\\
& SFT
& 38.67  & 13.60 & 37.23 & 12.76 & \textbf{38.67} & 6.03 & 36.77 & 6.90 & 13.60 & 6.03 & 12.76 & 6.90\\
\cdashlinelr{2-14}
& ST
& 35.78 & 13.76 & 33.51 & 13.61 & 37.22 & 8.62 & \textbf{37.91} & 8.62 & 13.79 & 14.45 & 12.41 & 6.90\\
& AL
& 41.77 & 24.66  & 39.42 & 23.30 & 37.30 & 18.86 & 36.39 & 15.52 & 23.13 & 20.69 & \textbf{23.64} & 23.27\\
& HCL
& \textbf{48.36}  & \textbf{28.06} & \textbf{47.76} & \textbf{28.57} & 38.28 & \textbf{20.68} & 36.92 & \textbf{22.41} & \textbf{25.85} & \textbf{23.27} & 22.62 & \textbf{28.44} \\

\midrule

\multirow{5}{*}{Mistral-7B} & Vanilla 
& 44.50  & 16.49 & 40.10 & 15.98 & 43.97 & 11.21 & 44.95 & 17.24 & 16.49 & 11.21 & 15.98 & 17.24\\
& SFT 
& 55.72  & 20.41 & 54.89 & 21.60 & 55.72 & 21.55 & \textbf{56.71} & 20.67 & 20.41 & 21.55 & 21.60 & 20.67 \\
\cdashlinelr{2-14}
& {ST}
& 55.72  & 22.45 & 55.34 & 22.11 & 57.16 & 25.00 & 54.97  & 28.45 & 22.28 & 25.00 & 22.62 & 18.97\\
& {AL}
& 56.63 & 32.65 & 55.95  & 32.31 & \textbf{57.69} & 37.93 & 54.97 & 28.49 & 34.01 &  38.79 & 33.33 & 33.62\\
& {HCL}
& \textbf{57.92}  & \textbf{36.56} & \textbf{57.01} & \textbf{35.88} & 57.39 & \textbf{40.51} & 53.98 & \textbf{31.03} & \textbf{35.54} & \textbf{44.82} &  \textbf{35.37} & \textbf{37.93}\\
        
\bottomrule

        \end{tabular}
    \caption{Accuracy(\%) of different LMs with different training strategies on MWP. ST, AL, HCL refer to skill training, applied learning and hierarchical curriculum learning respectively. \texttt{RAW} refers to testing on the origin GSM8k test set, \texttt{HARD} refers to testing on the augmented test set on specific atomic skill. In Mixture column, \texttt{ARITH} and \texttt{UNIT} refer to testing on the augmented test set on arithmetic and unit conversion skills respectively.}
    \label{tab:main}
\end{table*}

\paragraph{Unit Conversion Augmentation.} 
The main challenge in unit conversion lies in the diverse ways units are represented.
Statistical analysis of the data in GSM8K shown in Appendix~\ref{sec:ap_data_eval} shows that the representation of these units in GSM8K is quite uniform, leading to an incomplete evaluation of unit conversion skills.
Without altering the original meaning of the questions, we have diversified the representations of units within the same quantity type.
Tab.~\ref{tab:statistics-unitaug-metric} demonstrates that the enhanced data better tests unit conversion skills.

\subsection{Models and Baselines.}
We investigate the skill generalization on two models from different families, namely LLaMa-2 (7B; \citealp{touvron2023llama}) and Mistral (7B; \citealp{jiang2023mistral}).
The baselines we compared include the following two types: 
(1) \textbf{Vanilla} Model, which is a model without any special modifications or enhancements. 
(2) \textbf{SFT} Model, which has been supervised fine-tuned on the training set of MWP.
We test all the model with zero-shot and few-shot prompting.
The few-shot examples are drawn from the training set in applied learning.

\section{Experimental Analysis and Findings}

\subsection{\textit{RQ1: Can atomic skills generalize from prerequisite tasks to compositional tasks spontaneously?}}
\label{sec:rq1}

\paragraph{Atomic skill CANNOT generalize to compositional tasks spontaneously.}

Tab.~\ref{tab:main} illustrates the overall performance of different language models in compositional tasks.
We observe that language models do not get noticeable gain on the MWP after skill training.
LLaMA-2 only improve from 13.60\% to 13.76\% with zero-shot prompting in \texttt{HARD}, even experiencing a slight decrease on \texttt{RAW}.
This phenomenon is model-independent, so skill does not generalize from metrics perspective.
Moreover, as shown in Tab.~\ref{tab:casestudy}, ST model employs a format entirely distinct from that of the prerequisite task. 
This clearly indicates that atomic skills actually do not generalize from prerequisite tasks to complex tasks at all.

\definecolor{lightred}{RGB}{255,200,200}
\definecolor{lightgreen}{RGB}{152, 251, 152}

\begin{table}[t]
    \centering
    \small
    \begin{tabular}{ccc}
    \toprule
        \multirow{1}{*}{Method} & Response &  \\
    \midrule
\multirow{3}{*}{ST} & \multirow{3}{*}{\parbox{5.5cm}{\setlength{\fboxsep}{1pt}... The salesman sold 31 shoes for \colorbox{lightred}{31 * \$25 = 775}.Thus, the salesman made a profit of \colorbox{lightgreen}{775 - 340 = 435}. So the answer...}} & \multirow{3}{*}{\ding{55}}\\ \\ \\
        \midrule

\multirow{3}{*}{AL}  & \multirow{3}{*}{\parbox{5.5cm}{\setlength{\fboxsep}{1pt}The salesman sold 31 sneakers for \colorbox{lightgreen}{31 * \$25 = 31 * 25 = 31 * 20 + 31 * 5 = 620} \\ \colorbox{lightgreen}{+ 155 = 775} throughout the rest of the...}} & \multirow{3}{*}{\ding{51}} \\ \\ \\
\midrule
\multirow{3}{*}{HCL} &  \multirow{3}{*}{\parbox{5.5cm}{\setlength{\fboxsep}{1pt}The salesman sold 31 shoes for \$25 each, so his profit was \colorbox{lightgreen}{31 * \$25 = 31 * 25 = 31 * 20} \\ \colorbox{lightgreen}{+ 31 * 5 = 620 + 155 = 775}. In total ...}} &  \multirow{3}{*}{\ding{51}} \\ \\ \\
    \bottomrule
    \end{tabular}
    \caption{Example of LM's response to MWP with different training strategies. The last column indicates whether the prerequisite task format has been integrated.}
    \label{tab:casestudy}
\end{table}

\paragraph{Atomic skills can be induced to generalize through applied learning.}
HCL introduce the second phase of applied learning to induce skill generalization in hierarchical curriculum learning.
As seen in Tab.~\ref{tab:main}, LLaMA-2 significantly improve from 13.60\% to 28.76\% with zero-shot prompting in \texttt{HARD}, demonstrating the successful generalization of skills.
Case studies from Tab.~\ref{tab:casestudy} further show that models after applied learning (AL and HCL) are capable of integrating data from prerequisites into their responses to MWP, thus performing accurate calculations.
Therefore, although skills do not spontaneously generalize from prerequisite tasks to compositional tasks, they can be induced through applied learning.
Furthermore, we emphasize that this induced generalization needs to be achieved through training and cannot be replaced by few-shot prompting, as the metrics for few-shot learning do not surpass those for zero-shot learning.

\paragraph{The enhancement of compositional tasks stem from the improvement of atomic skills.}

\begin{figure}[t]
    \centering
    \includegraphics[width=0.48\textwidth]{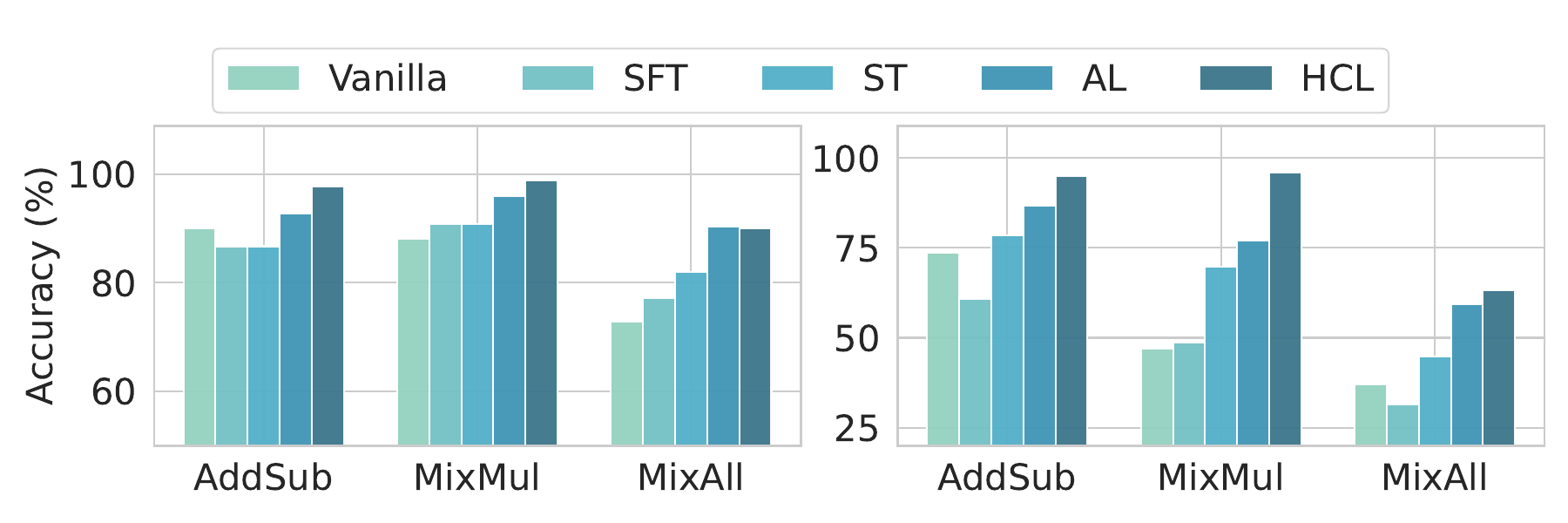}
    \caption{Accuracy(\%) of atomic skill on MWP of LLaMa-2. Left figure shows the results on \texttt{RAW} and right figure shows the results on \texttt{HARD}.}
    \label{fig:arith_acc}
\end{figure}

We extract the atomic skill part from the responses on MWP and calculate their accuracy, shown in Fig.~\ref{fig:arith_acc}.
Hierarchical curriculum learning results in significant improvements in arithmetic accuracy for all types of operations. 
The improvements are particularly striking for \texttt{MixMul} and \texttt{MixAll}, suggesting that current LMs are struggling to perform these arithmetic operations.
These improvement is consistent with the gain in answer accuracy for compositional tasks, as detailed in Appendix~\ref{sec:correlation}.

\begin{figure}[t]
    \centering
    \includegraphics[width=0.48\textwidth]{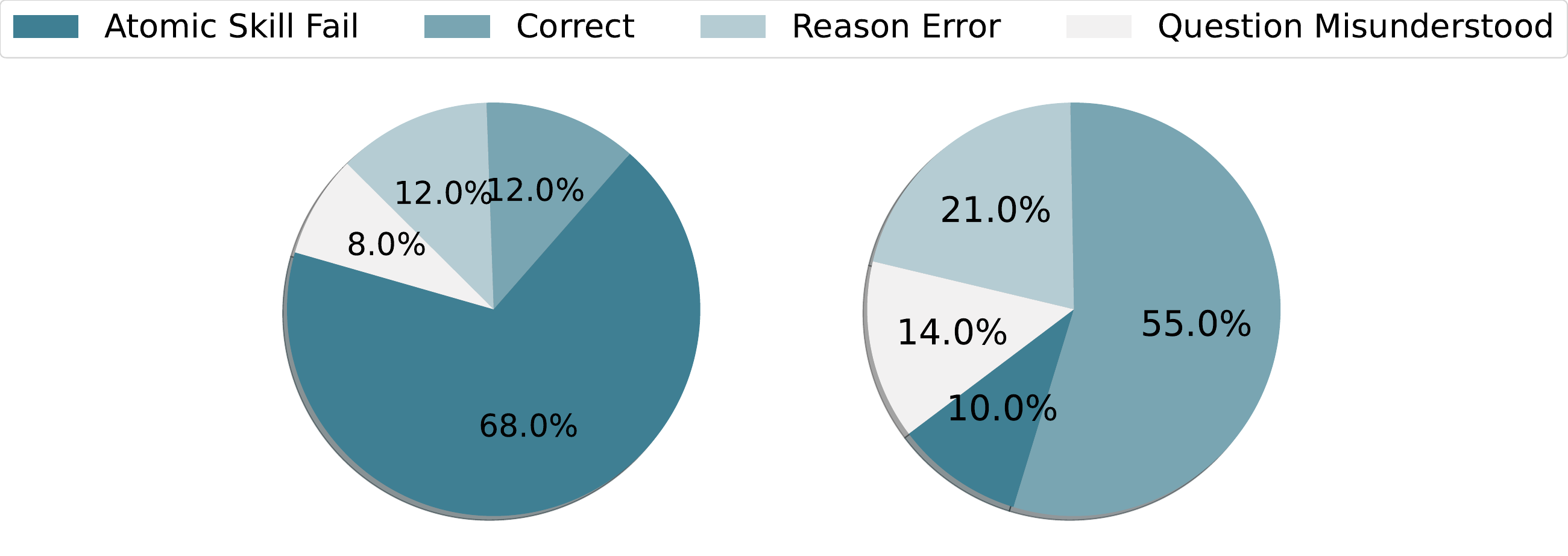}
    \caption{Error analysis on Vanilla model (left) and HCL model (right).}
    \label{fig:pie}
\end{figure}

Furthermore, we conduct an error analysis of the responses, as seen in Fig.~\ref{fig:pie}.
We first determine if a response involves an atomic skill error, and subsequently categorize other mistakes.
The majority of errors made by vanilla model stem from deficiencies in atomic skills.
After applying HCL, a few of errors shift to be question misunderstood and reasoning errors, while most errors are fully corrected to the right answers.
This demonstrates that the lack of atomic skills in a model \textit{masks} its superior reasoning capabilities, and HCL can effectively address this.

\paragraph{Mixture training is also effective for skill generalization.}
As seen in the third column of Tab.~\ref{tab:main}, mixture training yields similar results to achieve skill generalization as individual training. 
On the LLaMA-2 model with zero-shot prompting, arithmetic performance increase from 13.60\% to 25.85\%, and unit conversion performance increase from 6.03\% to 23.27\%.
Compared to individual training, mixture training leads to interactive effects, which depend on the intrinsic characteristics of the skills themselves.
From the results, the improvement in arithmetic from mixed training is not as significant as individual training. 
However, improvement in unit conversion was higher, possibly because enhanced arithmetic skills positively affected unit conversion, as accurate calculations aid in better conversions.

\begin{figure}[t]
    \centering
    \includegraphics[width=0.45\textwidth]{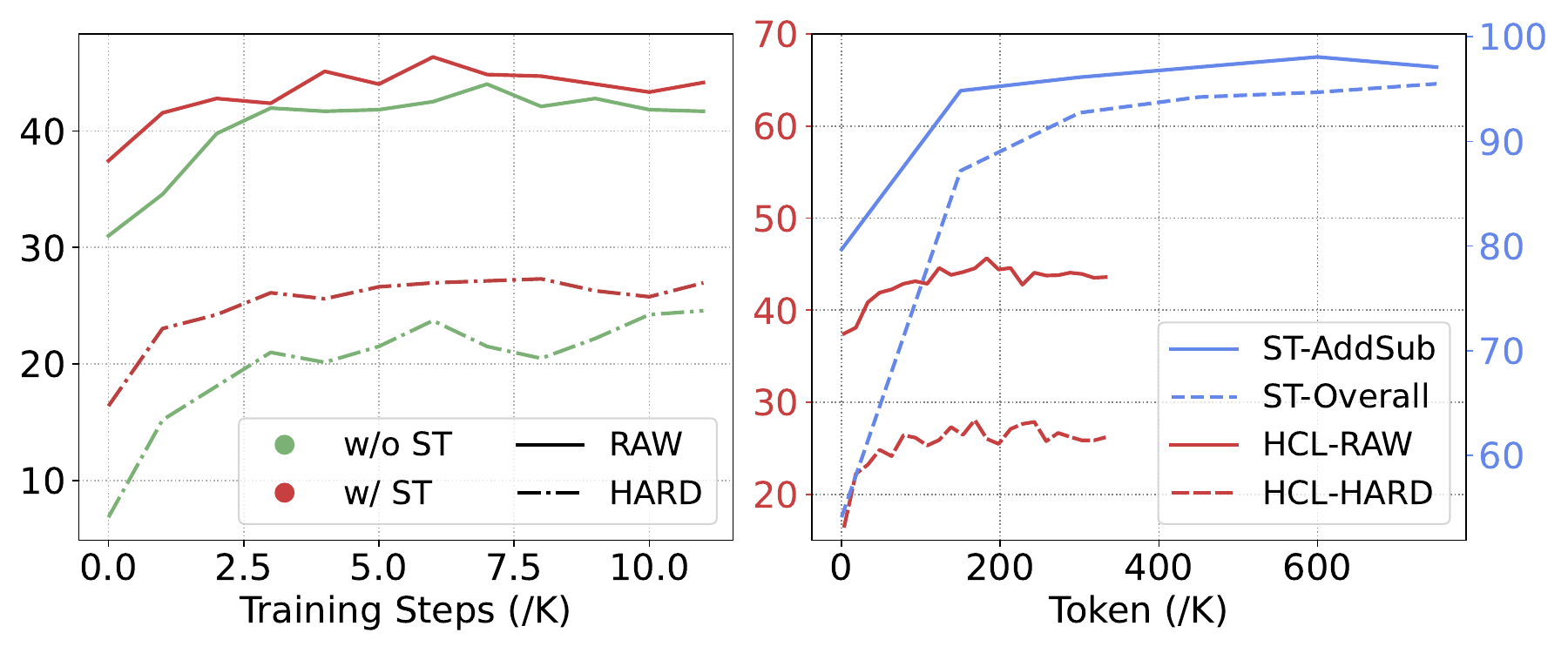}
    \caption{Accuracy(\%) of LLaMa-2 as training increases in different setting. 
    }
    \label{fig:step-token}
\end{figure}

\paragraph{Skill learning is indispensable in HCL.}
In Tab.~\ref{tab:main}, the performance of applied learning alone significantly lower than the full HCL.
Fig.~\ref{fig:step-token} left shows the accuracy of the applied learning and HCL with training increase and demonstrates that HCL reaches a better upper bound.
We argue that this is because applied learning only teaches the formal application of the skills without imparting the associated knowledge.

Furthermore, we observe that applied learning requires much less data compared to skill training.
In Fig.~\ref{fig:step-token} right, applied learning converge before 200K tokens, whereas skill training necessitates over 600K tokens.
This suggests that enhancing skills is more difficult than learning how to apply them. 
It underscores the educational principle that mastering prerequisite tasks is essential for solidifying atomic skills, while swift applied learning afterwards boosts the model's practical application capabilities.
In practical applications, it is often more challenging to obtain a large scale of heterogeneous compositional data than to acquire prerequisite data. 
The demand for data aligns with the challenges of data collection, which further demonstrates the feasibility of our approach.

\subsection{\textit{RQ2: Do atomic skills exhibits cross dataset or cross domain generalization?}}
Given that the data in applied learning is sourced from GSM8k, we aim to ascertain whether skill generalization is also effective in out-of-distribution (OOD) data.
We categorize the OOD data into two types: inter-domain and cross-domain. 
For inter-domain data, we use SVAMP~\cite{patel-etal-2021-nlp} and MathQA~\cite{amini-etal-2019-mathqa}, both of which comprise math word problems as well.
As for cross-domain data, we use the MMLU-Physics~\cite{hendrycks2020measuring}, as it is a physical task but also relies on arithmetic skill.

\definecolor{lightred}{RGB}{255,200,200}
\definecolor{lightgreen}{RGB}{152, 251, 152}

\begin{table}[t]
    \centering
    \small
    \begin{tabular}{cc}
    \toprule
        \multirow{1}{*}{Dataset} & Response  \\
    \midrule

\multirow{7}{*}{SVAMP} &  \multirow{3}{*}{\parbox{6cm}{\setlength{\fboxsep}{1pt}

\textbf{SFT}: They harvest 66 sacks per day and each sack contains 28 oranges so they harvest \colorbox{lightred}{66*28} \colorbox{lightred}{= 1808} oranges per day.

}} \\ \\ \\ 

\cdashlinelr{2-2}

 & \multirow{3}{*}{\parbox{6cm}{\setlength{\fboxsep}{1pt}

\textbf{HCL}: They harvest 66 sacks per day, and each sack contains 28 oranges, so they harvest \colorbox{lightgreen}{66 * 28 = 66 * 20 + 66 * 8 = 1320 + 528 = 1848} oranges per day. 
 
 }} \\ \\ \\ \\
        \midrule

\multirow{9}{*}{MathQA}  & \multirow{3}{*}{\parbox{6cm}{\setlength{\fboxsep}{1pt}

\textbf{SFT}: If selling an article for Rs. 1620 gives a 25\% profit, then the percentage profit earned is \colorbox{lightred}{25/100*1620 = 405}.

}} \\ \\ \\ 

\cdashlinelr{2-2}

  & \multirow{3}{*}{\parbox{6cm}{\setlength{\fboxsep}{1pt}

\textbf{HCL}: When the article is sold for Rs. 1620, the seller makes a profit of \colorbox{lightgreen}{25/100*1620 = 25 / 100} \colorbox{lightgreen}{* 1620 = 0.25 * 1620  = 0.25 * 1000 + 0.25 *600} \colorbox{lightgreen}{+ 0.25 * 20 + 0.25 * 0 = 250 + 150 + 5 + 0 = } \colorbox{lightgreen}{400 + 5 + 0 = 405 + 0 = 405.} When the article is sold for Rs. 1280, the ...

}} \\ \\ \\ \\ \\ \\
\midrule
\midrule
 &  \multirow{2}{*}{\parbox{6cm}{\setlength{\fboxsep}{1pt}
 
 \textbf{SFT}: The stone's speed in the air is 24 m/s * 9.8 m/s \textasciicircum 2 = \colorbox{lightred}{24 * 9.8 = 22.8}= 22.8 m/s. The ...
 
 }} \\  \\

\cdashlinelr{2-2}

 &  \multirow{3}{*}{\parbox{6cm}{\setlength{\fboxsep}{1pt}

 \textbf{HCL}: The horizontal component of the stone's speed is 24 m/s * 9.81 m/s\textasciicircum 2 = \colorbox{lightgreen}{24 * 9.81 = 24 *} \colorbox{lightgreen}{9 + 24 * 0.8 + 24 * 0.01 = 216 + 19.2 + 0.24 =} \colorbox{lightgreen}{235.2 + 0.24 = 235.44} = 235.44 m/s. The ...
 }} \\   \multirow{1}{*}{MMLU} \\  -Physics \\ \\

 \cmidrule{2-2}

 & \multirow{2}{*}{\parbox{6cm}{\setlength{\fboxsep}{1pt}

 \textbf{HCL}: ... Thus, R = PV/nT = \colorbox{orange}{(1.105 * 20 * 10\textasciicircum } \colorbox{orange}{-6) / (0.0451 * 273) = 8.314} J/mol*K. Since ...}} \\ \\
 
    \bottomrule
    \end{tabular}
    \caption{Example of response in cross-dataset and cross-domain senarios.}
    \label{tab:cross}
\end{table}

\paragraph{Skill generalization is effective in inter-domain data.}

As seen in Tab.~\ref{tab:cross}, with applied learning, models generates responses using step-wise format, unlike the original model which answered directly.
This demonstrates models can effectively utilize the skills even if the questions originating from a cross-dataset distribution.

\paragraph{Atomic skills can generalize across domains and show selective adaptability.}

In Tab.~\ref{tab:cross}, the model perform arithmetic among physical quantities in the same format as in prerequisite tasks, indicating that skill generalization still exhibits effectiveness in cross-domain scenarios.
LMs also exhibit selective adaptability when processing unseen data.
For instance, when dealing with the exponential value ``$10^{-6}$'' that is unseen in prerequisite tasks, LMs opt to answer in its original format.
Our findings reveal that while we need to introduce some compositional data during applied learning to induce skill generalization, it is not necessary to provide for every tasks.
However, it is crucial to have well-designed prerequisite data that can be applied across a broad spectrum of scenarios.

\subsection{\textit{RQ3: Can complex tasks help enhance atomic skills conversely?}}

\label{sec:rq4}

Considering that applied learning data itself combines multiple atomic skills, it suggests that compositional data may also have a positive effect on atomic skills.
We construct test dataset to assess the arithmetic skill, and then evaluate on the models with different training strategies.

\paragraph{Training with compositional data benefits the atomic skills, but the effect is limited.}

As shown in Tab.~\ref{tab:com2atomic}, applied learning achieve better performance in all operations compared to SFT, showing that compositional task have a positive effect on prerequisite data conversely.
This lead to a promising conclusion that training on complex reasoning dataset not only improve the performance of the specific task but also benefits its prerequisite tasks, as long as they exhibit \textit{composability} in response.
We further discuss the improvement in Appendix~\ref{sec:converse}.
Moreover, the gains brought by compositional tasks are minimal.
Applied learning only results in an increase from 22.91\% to 40.72\% on \texttt{C-Mul} while skill training leads to a skyrocketing increase to 93.45\%.
We attribute this to the limited heterogeneous data for applied learning.
The prerequisite tasks can automatically generate a large amount of heterogeneous data, but the compositional data is limited by the training set of the original training set for complex reasoning task.
Therefore, applied learning alone can not sufficiently enhance atomic skills, further highlighting the importance of a skill training stage in hierarchical curriculum learning.

\begin{table}[t]
    \centering
    \small
\resizebox{0.47\textwidth}{15mm} {
    \begin{tabular}{lcccccccc}
    \toprule
 \multirow{1}{*}{\textbf{Method}} & \multicolumn{2}{c}{\texttt{AddSub}} & \multicolumn{2}{c}{\texttt{S-Mul}} & \multicolumn{2}{c}{\texttt{C-Mul}} & \multicolumn{2}{c}{\texttt{S-Div}} \\
    \midrule
    
\rowcolor[gray]{0.95} \multicolumn{9}{c}{\textit{w/o Skill Training}}\\
        Vanilla
& \textbf{76.84} & & 51.65 & & 22.91  &  & 70.67 & \\
         AL
& 74.03 & \tiny(-2.81) & \textbf{52.32} & \tiny(+0.67) & \textbf{40.72} & \tiny(+17.81) & \textbf{78.94} & \tiny(+8.27) \\
\midrule
\rowcolor[gray]{0.95} \multicolumn{9}{c}{\textit{w/ Skill Training}}\\
        ST
& 97.54 &  & \textbf{95.75} &  & 93.45 & & 78.19 &  \\
        HCL
& \textbf{98.59} & \tiny(+1.05) & 82.27 & \tiny(-13.48) & \textbf{93.82} & \tiny(+0.37) & \textbf{81.20} & \tiny(+3.01)\\
    \bottomrule
    \end{tabular}}
    \caption{Accuracy(\%) in prerequisite tasks of LMs with and without skill training.}
    \label{tab:com2atomic}
\end{table}

\paragraph{Continued training on compositional tasks does not lead to catastrophic forgetting on atomic skills.}

Models with skill learning already possess proficient atomic skills but there is a risk that training on heterogeneous data in applied learning may lead to catastrophic forgetting.
However, as seen in Tab.~\ref{tab:com2atomic}, it striking is that HCL remains at a comparable level to skill training for most arithmetic operation.
This illustrates that continuous training on a compositional tasks, can spontaneously prevent catastrophic forgetting of atomic skills.
We suggest this due to the limited operations on compositional data in \texttt{S-Mul}, which may lead the model to be confused by other data within \texttt{C-Mul}. 
However, an anomaly is observed in \texttt{S-Mul} operation, which we further discuss in Appendix~\ref{sec:anomaly}.

\section{Conclusion}

In this work, we are the first to investigate the generalization from atomic tasks to complex reasoning tasks.
We propose a probing framework, in which we select math word problems as research example and arithmetic and unit conversion as related atomic skills.
By empirical experiments, we reveal that atomic skills can not spontaneously generalize to complex reasoning tasks.
Furthermore, we propose a hierarchical curriculum learning strategy to induce skill generalization and show effectiveness.
Our experimental findings provide valuable guidance for designing better training strategies for complex reasoning tasks in future work.

\section*{Limitations}
In this work, we choose math word problems as the research task , yet there are numerous more complicated reasoning tasks that rely on atomic skills, such as task planning, scenario modelling, decision making, and so on.
Although we do not delve deeply into more complicated and pluralistic reasoning tasks, these areas emerge a particularly interesting direction for future research.
Another limitation is that our proposed skill generalization depends on atomic skills that can be explicitly demonstrated in the response.
It remains uncertain whether implicit atomic skills can also have a positive effect on complex reasoning tasks.
Moreover, the definition of atomic skills and the method for prerequisite data generation are based on manual specification.
It is worth exploring automated methodologies to design a complete framework for hierarchical curriculum learning in future work.

\section*{Ethical Considerations}
All the data sources and language models used in this paper is available.
In this paper, most of the data generation and evaluation are automated, except for the error analysis in \S~\ref{sec:rq1} where human evaluation is used.
The details about human evaluation are provided in Appendix~\ref{sec:ap_human}. 
We protect the privacy rights of annotators. 
All annotators have been paid above the local minimum wage and consented to use the evaluation dataset for research purposes covered in our paper.
Our work does not raise any ethical considerations regarding potential risks and does not involve the research of human subjects.

\bibliography{reference}

\begin{thebibliography}{39}
\expandafter\ifx\csname natexlab\endcsname\relax\def\natexlab#1{#1}\fi

\bibitem[{Amini et~al.(2019)Amini, Gabriel, Lin, Koncel-Kedziorski, Choi, and Hajishirzi}]{amini-etal-2019-mathqa}
Aida Amini, Saadia Gabriel, Shanchuan Lin, Rik Koncel-Kedziorski, Yejin Choi, and Hannaneh Hajishirzi. 2019.
\newblock \href {https://doi.org/10.18653/v1/N19-1245} {{M}ath{QA}: Towards interpretable math word problem solving with operation-based formalisms}.
\newblock In \emph{Proceedings of the 2019 Conference of the North {A}merican Chapter of the Association for Computational Linguistics: Human Language Technologies, Volume 1 (Long and Short Papers)}, pages 2357--2367, Minneapolis, Minnesota. Association for Computational Linguistics.

\bibitem[{Bengio et~al.(2009)Bengio, Louradour, Collobert, and Weston}]{bengio2009curriculum}
Yoshua Bengio, J{\'e}r{\^o}me Louradour, Ronan Collobert, and Jason Weston. 2009.
\newblock Curriculum learning.
\newblock In \emph{Proceedings of the 26th annual international conference on machine learning}, pages 41--48.

\bibitem[{Chen et~al.(2023)Chen, Roberts, Bhatia, Wang, Zhang, Sala, and R{\'e}}]{Chen2023SkillitAD}
Mayee~F. Chen, Nicholas Roberts, Kush Bhatia, Jue Wang, Ce~Zhang, Frederic Sala, and Christopher R{\'e}. 2023.
\newblock \href {https://api.semanticscholar.org/CorpusID:260203057} {Skill-it! a data-driven skills framework for understanding and training language models}.
\newblock \emph{ArXiv}, abs/2307.14430.

\bibitem[{Cobbe et~al.(2021)Cobbe, Kosaraju, Bavarian, Chen, Jun, Kaiser, Plappert, Tworek, Hilton, Nakano, Hesse, and Schulman}]{cobbe2021gsm8k}
Karl Cobbe, Vineet Kosaraju, Mohammad Bavarian, Mark Chen, Heewoo Jun, Lukasz Kaiser, Matthias Plappert, Jerry Tworek, Jacob Hilton, Reiichiro Nakano, Christopher Hesse, and John Schulman. 2021.
\newblock Training verifiers to solve math word problems.
\newblock \emph{arXiv preprint arXiv:2110.14168}.

\bibitem[{Elgaar and Amiri(2023)}]{elgaar-amiri-2023-hucurl}
Mohamed Elgaar and Hadi Amiri. 2023.
\newblock \href {https://doi.org/10.18653/v1/2023.acl-long.104} {{H}u{C}url: Human-induced curriculum discovery}.
\newblock In \emph{Proceedings of the 61st Annual Meeting of the Association for Computational Linguistics (Volume 1: Long Papers)}, pages 1862--1877, Toronto, Canada. Association for Computational Linguistics.

\bibitem[{Hendrycks et~al.(2021)Hendrycks, Burns, Basart, Zou, Mazeika, Song, and Steinhardt}]{hendrycks2020measuring}
Dan Hendrycks, Collin Burns, Steven Basart, Andy Zou, Mantas Mazeika, Dawn Song, and Jacob Steinhardt. 2021.
\newblock \href {https://openreview.net/forum?id=d7KBjmI3GmQ} {Measuring massive multitask language understanding}.
\newblock In \emph{9th International Conference on Learning Representations, {ICLR} 2021, Virtual Event, Austria, May 3-7, 2021}. OpenReview.net.

\bibitem[{Huang and Chang(2023)}]{huang-chang-2023-towards}
Jie Huang and Kevin Chen-Chuan Chang. 2023.
\newblock \href {https://doi.org/10.18653/v1/2023.findings-acl.67} {Towards reasoning in large language models: A survey}.
\newblock In \emph{Findings of the Association for Computational Linguistics: ACL 2023}, pages 1049--1065, Toronto, Canada. Association for Computational Linguistics.

\bibitem[{Huang et~al.(2005)Huang, O'shaughnessy, and Wagner}]{huang2005prerequisite}
Jiunn Huang, John O'shaughnessy, and Robin Wagner. 2005.
\newblock Prerequisite change and its effect on intermediate accounting performance.
\newblock \emph{Journal of Education for Business}, 80(5):283--288.

\bibitem[{Huang et~al.(2023)Huang, He, Liang, Jiang, Xiao, and Chen}]{huang2023enhancing}
Yuncheng Huang, Qianyu He, Jiaqing Liang, Sihang Jiang, Yanghua Xiao, and Yunwen Chen. 2023.
\newblock Enhancing quantitative reasoning skills of large language models through dimension perception.
\newblock \emph{arXiv preprint arXiv:2312.17532}.

\bibitem[{Imani et~al.(2023)Imani, Du, and Shrivastava}]{imani-etal-2023-mathprompter}
Shima Imani, Liang Du, and Harsh Shrivastava. 2023.
\newblock \href {https://doi.org/10.18653/v1/2023.acl-industry.4} {{M}ath{P}rompter: Mathematical reasoning using large language models}.
\newblock In \emph{Proceedings of the 61st Annual Meeting of the Association for Computational Linguistics (Volume 5: Industry Track)}, pages 37--42, Toronto, Canada. Association for Computational Linguistics.

\bibitem[{Jiang et~al.(2023)Jiang, Sablayrolles, Mensch, Bamford, Chaplot, Casas, Bressand, Lengyel, Lample, Saulnier et~al.}]{jiang2023mistral}
Albert~Q Jiang, Alexandre Sablayrolles, Arthur Mensch, Chris Bamford, Devendra~Singh Chaplot, Diego de~las Casas, Florian Bressand, Gianna Lengyel, Guillaume Lample, Lucile Saulnier, et~al. 2023.
\newblock Mistral 7b.
\newblock \emph{arXiv preprint arXiv:2310.06825}.

\bibitem[{Jiang et~al.(2015)Jiang, Meng, Zhao, Shan, and Hauptmann}]{jiang2015self}
Lu~Jiang, Deyu Meng, Qian Zhao, Shiguang Shan, and Alexander Hauptmann. 2015.
\newblock Self-paced curriculum learning.
\newblock In \emph{Proceedings of the AAAI Conference on Artificial Intelligence}, volume~29.

\bibitem[{Ke and Liu(2022)}]{ke2022continual}
Zixuan Ke and Bing Liu. 2022.
\newblock Continual learning of natural language processing tasks: A survey.
\newblock \emph{arXiv preprint arXiv:2211.12701}.

\bibitem[{Keysers et~al.(2020)Keysers, Sch{\"{a}}rli, Scales, Buisman, Furrer, Kashubin, Momchev, Sinopalnikov, Stafiniak, Tihon, Tsarkov, Wang, van Zee, and Bousquet}]{keysers2019measuring}
Daniel Keysers, Nathanael Sch{\"{a}}rli, Nathan Scales, Hylke Buisman, Daniel Furrer, Sergii Kashubin, Nikola Momchev, Danila Sinopalnikov, Lukasz Stafiniak, Tibor Tihon, Dmitry Tsarkov, Xiao Wang, Marc van Zee, and Olivier Bousquet. 2020.
\newblock \href {https://openreview.net/forum?id=SygcCnNKwr} {Measuring compositional generalization: {A} comprehensive method on realistic data}.
\newblock In \emph{8th International Conference on Learning Representations, {ICLR} 2020, Addis Ababa, Ethiopia, April 26-30, 2020}. OpenReview.net.

\bibitem[{Khalifa et~al.(2023)Khalifa, Logeswaran, Lee, Lee, and Wang}]{grace2023}
Muhammad Khalifa, Lajanugen Logeswaran, Moontae Lee, Honglak Lee, and Lu~Wang. 2023.
\newblock \href {https://openreview.net/forum?id=2MiTZxLFA9} {Grace: Discriminator-guided chain-of-thought reasoning}.

\bibitem[{Khashabi et~al.(2020)Khashabi, Min, Khot, Sabharwal, Tafjord, Clark, and Hajishirzi}]{khashabi-etal-2020-unifiedqa}
Daniel Khashabi, Sewon Min, Tushar Khot, Ashish Sabharwal, Oyvind Tafjord, Peter Clark, and Hannaneh Hajishirzi. 2020.
\newblock \href {https://doi.org/10.18653/v1/2020.findings-emnlp.171} {{UNIFIEDQA}: Crossing format boundaries with a single {QA} system}.
\newblock In \emph{Findings of the Association for Computational Linguistics: EMNLP 2020}, pages 1896--1907, Online. Association for Computational Linguistics.

\bibitem[{Kim et~al.(2023)Kim, Asai, Ilharco, and Hajishirzi}]{kim-etal-2023-taskweb}
Joongwon Kim, Akari Asai, Gabriel Ilharco, and Hannaneh Hajishirzi. 2023.
\newblock \href {https://doi.org/10.18653/v1/2023.emnlp-main.680} {{T}ask{W}eb: Selecting better source tasks for multi-task {NLP}}.
\newblock In \emph{Proceedings of the 2023 Conference on Empirical Methods in Natural Language Processing}, pages 11032--11052, Singapore. Association for Computational Linguistics.

\bibitem[{Kim and Linzen(2020)}]{kim-linzen-2020-cogs}
Najoung Kim and Tal Linzen. 2020.
\newblock \href {https://doi.org/10.18653/v1/2020.emnlp-main.731} {{COGS}: A compositional generalization challenge based on semantic interpretation}.
\newblock In \emph{Proceedings of the 2020 Conference on Empirical Methods in Natural Language Processing (EMNLP)}, pages 9087--9105, Online. Association for Computational Linguistics.

\bibitem[{Lake and Baroni(2018)}]{lake2018generalization}
Brenden Lake and Marco Baroni. 2018.
\newblock Generalization without systematicity: On the compositional skills of sequence-to-sequence recurrent networks.
\newblock In \emph{International conference on machine learning}, pages 2873--2882. PMLR.

\bibitem[{Lewis et~al.(2020)Lewis, Perez, Piktus, Petroni, Karpukhin, Goyal, K{\"u}ttler, Lewis, Yih, Rockt{\"a}schel et~al.}]{lewis2020retrieval}
Patrick Lewis, Ethan Perez, Aleksandra Piktus, Fabio Petroni, Vladimir Karpukhin, Naman Goyal, Heinrich K{\"u}ttler, Mike Lewis, Wen-tau Yih, Tim Rockt{\"a}schel, et~al. 2020.
\newblock Retrieval-augmented generation for knowledge-intensive nlp tasks.
\newblock \emph{Advances in Neural Information Processing Systems}, 33:9459--9474.

\bibitem[{Liu and Low(2023)}]{liu2023goat}
Tiedong Liu and Bryan Kian~Hsiang Low. 2023.
\newblock Goat: Fine-tuned llama outperforms gpt-4 on arithmetic tasks.
\newblock \emph{arXiv preprint arXiv:2305.14201}.

\bibitem[{McCloskey and Cohen(1989)}]{mccloskey1989catastrophic}
Michael McCloskey and Neal~J Cohen. 1989.
\newblock Catastrophic interference in connectionist networks: The sequential learning problem.
\newblock In \emph{Psychology of learning and motivation}, volume~24, pages 109--165. Elsevier.

\bibitem[{Muffo et~al.(2022)Muffo, Cocco, and Bertino}]{muffo-etal-2022-evaluating}
Matteo Muffo, Aldo Cocco, and Enrico Bertino. 2022.
\newblock \href {https://aclanthology.org/2022.lrec-1.30} {Evaluating transformer language models on arithmetic operations using number decomposition}.
\newblock In \emph{Proceedings of the Thirteenth Language Resources and Evaluation Conference}, pages 291--297, Marseille, France. European Language Resources Association.

\bibitem[{Nogueira et~al.(2021)Nogueira, Jiang, and Lin}]{nogueira2021investigating}
Rodrigo Nogueira, Zhiying Jiang, and Jimmy Lin. 2021.
\newblock Investigating the limitations of transformers with simple arithmetic tasks.
\newblock \emph{arXiv preprint arXiv:2102.13019}.

\bibitem[{Park et~al.(2022)Park, Ryu, and Choi}]{park-etal-2022-language}
Sungjin Park, Seungwoo Ryu, and Edward Choi. 2022.
\newblock \href {https://doi.org/10.18653/v1/2022.findings-emnlp.128} {Do language models understand measurements?}
\newblock In \emph{Findings of the Association for Computational Linguistics: EMNLP 2022}, pages 1782--1792, Abu Dhabi, United Arab Emirates. Association for Computational Linguistics.

\bibitem[{Patel et~al.(2021)Patel, Bhattamishra, and Goyal}]{patel-etal-2021-nlp}
Arkil Patel, Satwik Bhattamishra, and Navin Goyal. 2021.
\newblock \href {https://doi.org/10.18653/v1/2021.naacl-main.168} {Are {NLP} models really able to solve simple math word problems?}
\newblock In \emph{Proceedings of the 2021 Conference of the North American Chapter of the Association for Computational Linguistics: Human Language Technologies}, pages 2080--2094, Online. Association for Computational Linguistics.

\bibitem[{Rovick et~al.(1999)Rovick, Michael, Modell, Bruce, Horwitz, Adamson, Richardson, Silverthorn, and Whitescarver}]{rovick1999accurate}
Allen~A Rovick, Joel~A Michael, Harold~I Modell, David~S Bruce, Barbara Horwitz, Thomas Adamson, Daniel~R Richardson, Dee~U Silverthorn, and Shirley~A Whitescarver. 1999.
\newblock How accurate are our assumptions about our students' background knowledge?
\newblock \emph{Advances in Physiology Education}, 276(6):S93.

\bibitem[{Sanh et~al.(2022)Sanh, Webson, Raffel, Bach, Sutawika, Alyafeai, Chaffin, Stiegler, Raja, Dey, Bari, Xu, Thakker, Sharma, Szczechla, Kim, Chhablani, Nayak, Datta, Chang, Jiang, Wang, Manica, Shen, Yong, Pandey, Bawden, Wang, Neeraj, Rozen, Sharma, Santilli, F{\'{e}}vry, Fries, Teehan, Scao, Biderman, Gao, Wolf, and Rush}]{sanh2021multitask}
Victor Sanh, Albert Webson, Colin Raffel, Stephen~H. Bach, Lintang Sutawika, Zaid Alyafeai, Antoine Chaffin, Arnaud Stiegler, Arun Raja, Manan Dey, M~Saiful Bari, Canwen Xu, Urmish Thakker, Shanya~Sharma Sharma, Eliza Szczechla, Taewoon Kim, Gunjan Chhablani, Nihal~V. Nayak, Debajyoti Datta, Jonathan Chang, Mike~Tian{-}Jian Jiang, Han Wang, Matteo Manica, Sheng Shen, Zheng~Xin Yong, Harshit Pandey, Rachel Bawden, Thomas Wang, Trishala Neeraj, Jos Rozen, Abheesht Sharma, Andrea Santilli, Thibault F{\'{e}}vry, Jason~Alan Fries, Ryan Teehan, Teven~Le Scao, Stella Biderman, Leo Gao, Thomas Wolf, and Alexander~M. Rush. 2022.
\newblock \href {https://openreview.net/forum?id=9Vrb9D0WI4} {Multitask prompted training enables zero-shot task generalization}.
\newblock In \emph{The Tenth International Conference on Learning Representations, {ICLR} 2022, Virtual Event, April 25-29, 2022}. OpenReview.net.

\bibitem[{Schick et~al.(2023)Schick, Dwivedi-Yu, Dess{\`\i}, Raileanu, Lomeli, Zettlemoyer, Cancedda, and Scialom}]{schick2023toolformer}
Timo Schick, Jane Dwivedi-Yu, Roberto Dess{\`\i}, Roberta Raileanu, Maria Lomeli, Luke Zettlemoyer, Nicola Cancedda, and Thomas Scialom. 2023.
\newblock Toolformer: Language models can teach themselves to use tools.
\newblock \emph{arXiv preprint arXiv:2302.04761}.

\bibitem[{Scott(2008)}]{2008STUDENT}
Shaun~Eric Scott. 2008.
\newblock Student academic performance in skills-based technology courses delivered through different scheduling formats.
\newblock \emph{Dissertations \& Theses - Gradworks}.

\bibitem[{Talmor and Berant(2019)}]{talmor-berant-2019-multiqa}
Alon Talmor and Jonathan Berant. 2019.
\newblock \href {https://doi.org/10.18653/v1/P19-1485} {{M}ulti{QA}: An empirical investigation of generalization and transfer in reading comprehension}.
\newblock In \emph{Proceedings of the 57th Annual Meeting of the Association for Computational Linguistics}, pages 4911--4921, Florence, Italy. Association for Computational Linguistics.

\bibitem[{Taori et~al.(2023)Taori, Gulrajani, Zhang, Dubois, Li, Guestrin, Liang, and Hashimoto}]{alpaca}
Rohan Taori, Ishaan Gulrajani, Tianyi Zhang, Yann Dubois, Xuechen Li, Carlos Guestrin, Percy Liang, and Tatsunori~B. Hashimoto. 2023.
\newblock Stanford alpaca: An instruction-following llama model.
\newblock \url{https://github.com/tatsu-lab/stanford_alpaca}.

\bibitem[{Touvron et~al.(2023)Touvron, Martin, Stone, Albert, Almahairi, Babaei, Bashlykov, Batra, Bhargava, Bhosale et~al.}]{touvron2023llama}
Hugo Touvron, Louis Martin, Kevin Stone, Peter Albert, Amjad Almahairi, Yasmine Babaei, Nikolay Bashlykov, Soumya Batra, Prajjwal Bhargava, Shruti Bhosale, et~al. 2023.
\newblock Llama 2: Open foundation and fine-tuned chat models.
\newblock \emph{arXiv preprint arXiv:2307.09288}.

\bibitem[{Wei et~al.(2022{\natexlab{a}})Wei, Bosma, Zhao, Guu, Yu, Lester, Du, Dai, and Le}]{wei2021finetuned}
Jason Wei, Maarten Bosma, Vincent~Y. Zhao, Kelvin Guu, Adams~Wei Yu, Brian Lester, Nan Du, Andrew~M. Dai, and Quoc~V. Le. 2022{\natexlab{a}}.
\newblock \href {https://openreview.net/forum?id=gEZrGCozdqR} {Finetuned language models are zero-shot learners}.
\newblock In \emph{The Tenth International Conference on Learning Representations, {ICLR} 2022, Virtual Event, April 25-29, 2022}. OpenReview.net.

\bibitem[{Wei et~al.(2022{\natexlab{b}})Wei, Wang, Schuurmans, Bosma, Xia, Chi, Le, Zhou et~al.}]{wei2022chain}
Jason Wei, Xuezhi Wang, Dale Schuurmans, Maarten Bosma, Fei Xia, Ed~Chi, Quoc~V Le, Denny Zhou, et~al. 2022{\natexlab{b}}.
\newblock Chain-of-thought prompting elicits reasoning in large language models.
\newblock \emph{Advances in Neural Information Processing Systems}, 35:24824--24837.

\bibitem[{White and Gagn{\'e}(1974)}]{white1974past}
Richard~T White and Robert~M Gagn{\'e}. 1974.
\newblock Past and future research on learning hierarchies.
\newblock \emph{Educational psychologist}, 11(1):19--28.

\bibitem[{Wu et~al.(2021)Wu, Dyer, and Neyshabur}]{wu2020curricula}
Xiaoxia Wu, Ethan Dyer, and Behnam Neyshabur. 2021.
\newblock \href {https://openreview.net/forum?id=tW4QEInpni} {When do curricula work?}
\newblock In \emph{9th International Conference on Learning Representations, {ICLR} 2021, Virtual Event, Austria, May 3-7, 2021}. OpenReview.net.

\bibitem[{Xu et~al.(2020)Xu, Zhang, Mao, Wang, Xie, and Zhang}]{xu-etal-2020-curriculum}
Benfeng Xu, Licheng Zhang, Zhendong Mao, Quan Wang, Hongtao Xie, and Yongdong Zhang. 2020.
\newblock \href {https://doi.org/10.18653/v1/2020.acl-main.542} {Curriculum learning for natural language understanding}.
\newblock In \emph{Proceedings of the 58th Annual Meeting of the Association for Computational Linguistics}, pages 6095--6104, Online. Association for Computational Linguistics.

\bibitem[{Ye et~al.(2021)Ye, Lin, and Ren}]{ye-etal-2021-crossfit}
Qinyuan Ye, Bill~Yuchen Lin, and Xiang Ren. 2021.
\newblock \href {https://doi.org/10.18653/v1/2021.emnlp-main.572} {{C}ross{F}it: A few-shot learning challenge for cross-task generalization in {NLP}}.
\newblock pages 7163--7189.

\end{thebibliography}

\clearpage

\appendix
\section{Detail for Training Data}
\label{sec:ap_data_construct}

\subsection{Arithmetic Prerequisite Task}
\label{sec:ap_arith_data}

We construct data for arithmetic prerequisite task by a rule-based approach, seen in Algorithm~\ref{algorithm:arith_datagen}.
The difficulty of the data considers four aspects: the number of hops, the length of a significant digit, the type of value and the type of operation.
The number of hops ranges from two to five, the more hops means the harder it is.
The significant digit length ranges from one to eight. 
Longer length means the more complex and difficult for calculations.
Value types include: all integers, all floating, and mixed data types, with increasing difficulty.
Operation types include \texttt{AddSub}, \texttt{W-Mul}, \texttt{C-Mul} and \texttt{S-Div}.
Among them, \texttt{AddSub} consists of only addition and subtraction, \texttt{S-Mul} involves simple multiplication operations (the second value with a significant figure of 1), \texttt{C-Mul} encompasses complex multiplication operations where we break down the second number and perform step-wise calculations, and \texttt{S-Div} represents simple division.

\begin{algorithm}[h]
    \footnotesize
    \SetAlgoLined
    \caption{Arithmetic Data Generation}
    \label{algorithm:arith_datagen}
    \KwData{Operation set $O$, Significant Digit set $D$, Value type set $V$, Hop set $H$}
    \KwResult{Arithmetic Expression $E$, Response $R$}
    
    \SetKwFunction{RG}{ResponseGeneration}
    \SetKwFunction{OHG}{OneHopResponse}
    
    \tcp{Initialization}
    $h \gets \text{Random}(H)$;\ \  $o \gets \text{Random}(O)$\;
    $d \gets \text{Random}(D)$;\ \  $v \gets \text{Random}(V)$\;
    $num_0 \gets \text{NumberGenerator}(d, v)$; 
    $E \gets num_0$;
    \tcp{Expression generation}
    \For{$i \gets 1$ \KwTo $h$}{
        $op \gets \text{OperationGenerator}(o)$\;
        $n_{i+1} \gets \text{NumberGenerator}(d, v, op)$\; 
        $E \gets E \; \circ \; op \; \circ \; n_{i+1}$\; 
    }
    $R \gets $ \RG($E$)\;
    \Return{$E$, $R$};

    \SetKwProg{Fn}{Function}{:}{}
    \tcp{Iterative response generation}
    \Fn{\RG{$exp$}}{
        $i \gets idx$ if exists $op_{idx}$ in $MulDiv$ else $0$\;
        $E_p \gets (n_0, op_0, \ldots, op_{i-1}, n_i)$\; 
        $E_s \gets (op_{i+1}, \ldots, n_{n})$\;
        $SR \gets \OHG(n_i, op_i, n_{i+1})$\; 
        $n_{new} \gets \text{eval}(n_i, op_i, n_{i+1})$\;
        $MR \gets \text{MergeResponse}(E_p, SR, E_s)$\; 
        $E \gets E_p \circ n_{new} \circ E_s$\;
        \KwRet MR + \RG(E); 
    }
    \tcp{One Hop COT response generation}
    \Fn{\OHG{$n_0$, $op$, $n_1$}}{
        \uIf{$op = \text{``C-MUL''}$}{
            $w_0, w_1, ..., w_m \gets \text{SplitDigits}(n_1)$\;
            $SR \gets \text{SplitCOT}(n_0, w_0, w_1, ..., w_m)$\;
            $SR$ += $\text{MulCOT}(n_0, w_0, w_1, ..., w_m)$\;
            $SR$ += $\text{AddCOT}(n_0 * w0, ... , n_0 * w_m)$\;
        }
        \Else{
            $SR \gets \text{Eval}(n_0, op, n_1)$\;
        }
        \KwRet $SR$\;
  }
\end{algorithm}

In Algorithm~\ref{algorithm:arith_datagen}, reponse demonstrates a step-wise arithmetic process based on hop.
For \texttt{C-Mul} operation, the second value is split and then perform bit-wise multiplication and results addition.
SplitCOT, MulCOT and AddCOT represent the process of digits splitting, bit-wise multiplication and results addition respectively.

\subsection{Unit Conversion Prerequisite Task}
\label{sec:ap_unit_data}

We construct data for unit conversion prerequisite task based on DimUnitKB~\cite{huang2023enhancing}.
We first extract the quantity types of units contained in the MWP, including seven type in total: length, time, speed, mass, volume, area and power.
 We construct pair-wise conversion data for units representing the same quantity types, detailed in Algorithm~\ref{algorithm:unit_datagen}.

\begin{algorithm}[h]
    \footnotesize
    \SetAlgoLined
    \caption{UnitConv. Data Generation}
    \label{algorithm:unit_datagen}
    \KwData{Quantity Type Set $Q$, DimUnitKB $K$}
    \KwResult{Unit Conversion Text Data $T$}

   \tcp{Select random quantity type and units}
    $q \leftarrow \text{Random}(S)$\;
    $U_q \leftarrow \{u \in K \mid u.\text{type} = q\}$\;
    $u_0, u_1 \leftarrow \text{Random}(U_q, 2)$\;
    \tcp{Calculate conversion ratio}
    $conv \leftarrow u_0.\text{conv} / u_1.\text{conv}$\;
    \tcp{Generate conversion text}
    $T \leftarrow$ TextGeneration($u_0$, $u_1$, $conv$)\;
    \Return{$T$};
\end{algorithm}

\subsection{Compositional Data}
\label{sec:ap_composition}
In applied learning, we need to construct data that integrates atomic skills into compositional tasks. 
The construction of compositional data consists of the following steps:
\begin{enumerate}
    \setlength{\itemsep}{0pt}
    \setlength{\parsep}{0pt}
    \setlength{\topsep}{0pt}
    \item Sample data items from the train set of complex reasoning tasks.
    \item Extract all segments related to atomic skills from the response of the item.
    \item Use the \texttt{ResponseGeneration} function and \texttt{TextGeneration} function in Algorithm~\ref{algorithm:arith_datagen} and Algorithm~\ref{algorithm:unit_datagen} to generate the new format of response for atomic skill.
    \item Replace all segments related to atomic skills with the new answer response to construct compositional data.
\end{enumerate}

For example, in arithmetic, we extract the arithmetic segments ``12 * 37 = 448'' from the response ``the total length of the ribbon is: 12 * 37 = 448 cm'' and then replace it  with ``12 * 37 = 12 * (30 + 7) = 12 * 30 + 12 * 7 = 360 + 84 = 444'' to get the compositional data ``the total length of the ribbon is: 12 * 37 = 12 * (30 + 7) = 12 * 30 + 12 * 7 = 360 + 84 = 444 cm''.

\section{Detail for Evaluation Data}
\label{sec:ap_data_eval}

\subsection{Deficiency of GSM8K in Skill Assessment}

GSM8K~\cite{cobbe2021gsm8k} is the current widely used evaluation benchmark for math word problem.
However, it does not comprehensively cover the application of atomic skills across various difficulty levels.
We analyze the difficulty coverage of GSM8k on arithmetic, taking into account four aspects mentioned in Section~\ref{sec:ap_arith_data}, as depicted in Fig.~\ref{fig:arith_statistic} (left).
Darker colours indicate more difficult, and a greater area signifies a larger volume of data.
It shows that the GSM8k test set comprehensively covers various operation hops, operation types, and value types. 
However, the significant digits involved are primarily focused on simple data.
GSM8k dataset also lacks comprehensiveness on unit conversion skills. 
The statistic is shown in Tab.~\ref{tab:statistics-unitaug}.
It covers only a few unit expressions in various types.

\begin{table}[t]
    \centering
    \small
    \begin{tabular}{ccccc}
    \toprule
        \multirow{1}{*}{Dataset} & \# w/ unit conv. & \# Length & \# Mass & \# Power \\
    \midrule
        \multirow{1}{*}{GSM8k$_\texttt{RAW}$}
& 61 & 3 & 5 & 2 \\
        \midrule
        \multirow{1}{*}{GSM8k$_\texttt{HARD}$}
& 116 & 10 & 13 & 4 \\
    \bottomrule
    \end{tabular}
    \caption{Partial statistic of \texttt{RAW} and \texttt{HARD} set.}
    \label{tab:statistics-unitaug}
\end{table}

\begin{algorithm}[t]
    \footnotesize
    \SetAlgoLined
    \caption{Augmenting MWP on Arith.}
    \label{algorithm:arith_dataaug}
    \KwIn{Original math problem $(Q_{RAW}, A_{RAW})$}
    \KwOut{Enhanced math problem $(Q_{HARD}, A_{HARD})$}
    
    $N \leftarrow \text{ExtractNumber}(Q_{RAW})$\;
    $I \leftarrow \text{ExtractNumber}(A_{RAW}) \setminus N$\;
    \tcp{Determine the computational relationship for $I$}
    $f \leftarrow \text{ExtractMapping}(A_{RAW})$\;
    $D \gets$ Random a maximum significant digit length\;
    \tcp{Generate new numbers for N with length <= D}
    \For{$i \leftarrow 1$ \KwTo $\text{Length}(N)$}{
        $N[i] \leftarrow \text{RandomNumber}(D)$\;
    }
    \tcp{Compute new values for I based on f and updated N}
    $I \leftarrow f(N, I)$\;
    \tcp{Substitute the new numbers into the original problem}
    $Q_{HARD}, A_{HARD} \leftarrow \text{Substitute}(Q_{RAW}, A_{RAW}, N, I)$\;

    \Return{$Q_{HARD}, A_{HARD}$};
    
\end{algorithm}

\begin{table}[t]
    \centering
    \small
    \begin{tabular}{ccccc}
    \toprule
        \multirow{2}{*}{\textbf{Model}} & \multicolumn{2}{c}{\textbf{GSM8k\tiny{\texttt{RAW}}}} & \multicolumn{2}{c}{\textbf{GSM8k\tiny{\texttt{HARD}}}} \\
    \cmidrule{2-3} \cmidrule{4-5}
        & overall & w/ unit conv. & overall & w/ unit conv. \\
    \midrule
        \multirow{1}{*}{LLaMa-2-7B}
& 27.16 & 21.31 & 14.44 & 1.72 \\
        \midrule
        \multirow{1}{*}{Mistral-7B}
& 45.10 & 47.54 & 29.46 & 13.79 \\
    \bottomrule
    \end{tabular}
    \caption{Performance of LLaMa-2 on \texttt{RAW} and \texttt{HARD} dataset in unit conversion skill.}
    \label{tab:statistics-unitaug-metric}
\end{table}

\subsection{Method for Data Augmenting}

As for arithmetic, we augment the evaluation data through increasing the significant digit lengths without changing the logic of the original problems and denote as \texttt{HARD} set.
The algorithm is outlined in Algorithm~\ref{algorithm:arith_dataaug}. 
Initially, we extract the numbers in the question, along with the intermediate numbers in the answer, and the computation relationships between them. 
Subsequently, we randomly get new numbers based on the maximum significant value length, and compute new intermediate numbers accordingly. 
These updated numbers are then integrated into the original question and answer, resulting in an enhanced question-answer pairs.
As shown in Fig.~\ref{fig:arith_statistic} (right), the \texttt{HARD} set covers a broader range of significant digits.
The new data distribution is reasonable with the proportion decreasing appropriately as the difficulty increases.

\begin{algorithm}[t]
    \footnotesize
    \SetAlgoLined
    \caption{Augmenting MWP on Unit Conversion.}
    \label{algorithm:unit_dataaug}
    \KwIn{Original math problem $(Q_{RAW}, A_{RAW})$, DimUnitKB $K$}
    \KwOut{Augmented problem $(Q_{HARD}, A_{HARD})$}
    
    $u_0 \leftarrow \text{ExtractTargetUnit}(Q_{RAW}, K)$\;
    $q \leftarrow \text{GetQuantityKind}(u_0, K)$\;
    \tcp{Randomly select a new unit with the same quantity type}
    $U_1 \leftarrow \text{RandomSelect}(\{u \in K \mid u.\text{type} = q\})$\;
    $conv \leftarrow u_0.\text{conv} / u_1.\text{conv}$\;
    $(Q_{HARD}, A_{HARD}) \leftarrow \text{Substitude}(Q_{RAW}, A_{RAW}, u_0, u_1, \text{conv})$\;

    \Return{$Q_{HARD}, A_{HARD}$};
    
\end{algorithm}

As for unit conversion, we augment the test data by including a wider variety of unit representations.
The algorithm is outlined in Algorithm~\ref{algorithm:arith_dataaug}. 
We evaluate LLAMA-2 on both \texttt{RAW} and \texttt{HARD}. As shown in Table~\ref{tab:statistics-unitaug-metric}, the results suggest that the \texttt{HARD} set provides greater differentiation.

\section{Experimental Details}

The implementations of all the LMs in our paper are based on the HuggingFace Transformers\footnote{https://github.com/huggingface/transformers/} and Deepspeed\footnote{https://github.com/microsoft/DeepSpeed}.
We set the learning rate in {1e-5, 1e-6} with a WarmupLR scheduler, batch size of 32, max sequence length of 1024 and train for 8 epochs.
All of our experiments are conducted on the workstations of NVIDIA A800 PCIe with 80GB memory and the environment of Ubuntu 20.04.6 LTS and torch 2.0.1.
In the evaluation, we employ vllm~\footnote{https://github.com/vllm-project/vllm} for inference. All the result are generate with a greedy decoding strategy.

We do not utilize additional prompts for prerequisite data but employ the prompt from Alpaca~\citep{alpaca} for training and testing on MWP.

\subsection{The Ratio Selection for Replay Strategy}
\label{sec:ratio}

\begin{figure}[h]
    \centering
    \includegraphics[width=0.48\textwidth, height=0.16\textwidth]{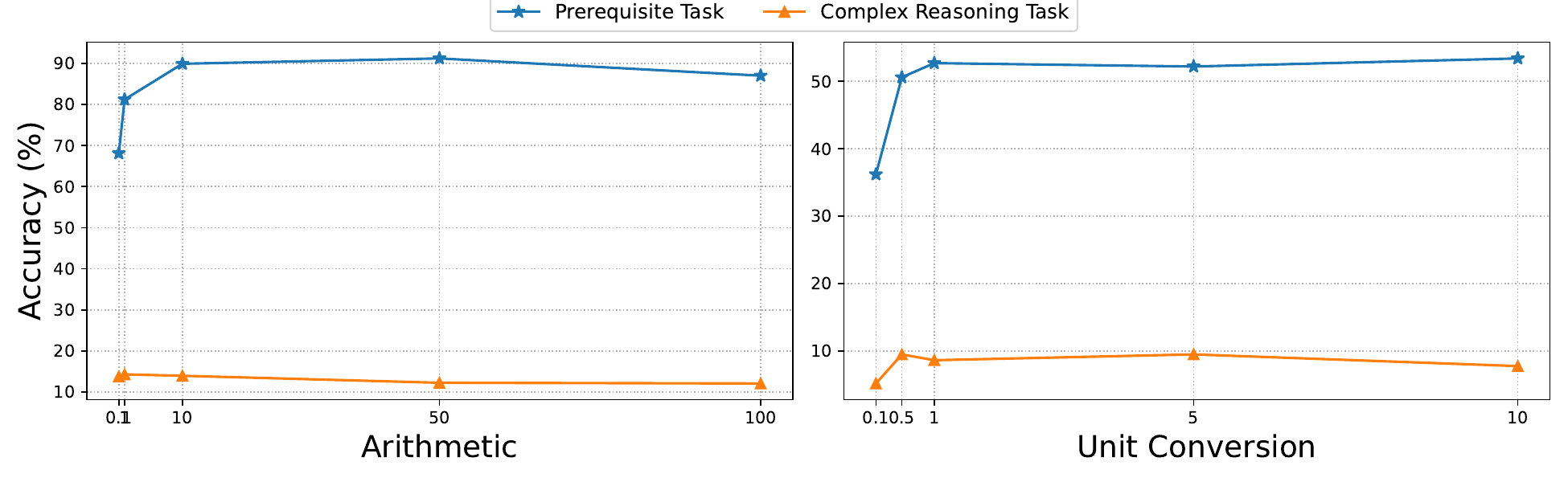}
    \caption{Accuracy of LMs on prerequisite tasks and complex reasoning tasks with different mixing ratio.}
    \label{fig:ratio}
\end{figure}

We retain some training examples from MWP and mix them with prerequisite task data to ensure the model retains its original problem-solving abilities in skill training.
Fig.~\ref{fig:ratio} shows the performance of the model at different mixing ratios on prerequisite tasks (atomic skills) and complex reasoning tasks. 

Overall, as the proportion of prerequisite data increases, the model become better in atomic skills and worse in solving MWP. 
The balanced mixing ratios vary for different skills.
Arithmetic requires more prerequisite data than unit conversion.
Ultimately, we chose to set the mixing ratio for arithmetic to 10 and for unit conversion to 1 in skill training, to achieve a relatively effective continual learning.

\subsection{Human Evaluation for Error Analysis}
\label{sec:ap_human}

We recruite human evaluators to do error analysis on math word problem.
All evaluators possess sufficient knowledge of mathematics and are provided with the necessary background for the evaluation criteria.
Each item is annotated by at least three evaluators and inconsistencies will lead to a reassessment.

The evaluation of the response is divided into four types:  1.Atomic skill error: an error occurs within the atomic skill segment, e.g., an error in performing complex multiplication; 2.Question Misunderstood: the wrong value is used due to a misunderstanding of the question; and 3.Reason error: incorrect representation due to an error in the reasoning process. 4.Correct: indicating that both the reasoning process and the answer are completely correct;
The priority of these four types of classifications decreases in order.

\section{Additional Results}

\subsection{The correlation of atomic skills}
\label{sec:correlation}

\begin{figure}[h]
    \centering
    \includegraphics[width=0.5\textwidth, height=0.3\textwidth]{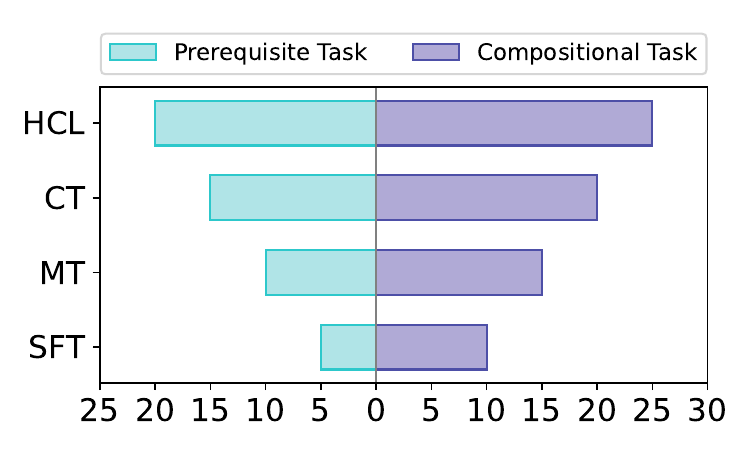}
    \caption{Accuracy gain (\%) on prerequisite tasks and composisiton tasks.}
    \label{fig:correlation}
\end{figure}

Fig.~\ref{fig:correlation} illustrates the correlation between the accuracy gains in prerequisite tasks and compositional tasks. 
A higher performance on prerequisite tasks indicates a stronger atomic skill of the model, revealing a positive correlation between the model's atomic skills and its capability to solve compositional tasks. 
This supports our conclusion that the gains brought by our method are attributable to the enhancement of atomic skills.

\subsection{Compositional task is not enough to enhanced atomic skills}
\label{sec:converse}

We mention in Section~\ref{sec:rq4} that the compositional task improves atomic skills, but the effect is limited.
It can be seen from Tab.~\ref{tab:conv_example} that although the AL method learn the COT form to answer, it has not acquired complete knowledge, hence it cannot result in a completely correct answer.

\subsection{Abnormal Results in \texttt{S-Mul}}
\label{sec:anomaly}
We mention in Section~\ref{sec:rq4} that HCL exhibit abnormal decline in \texttt{S-Mul} compared to ST.
We observe that most of the error comes from the incorrect of magnitude.
For example, HCL model reponse in ``0.00261'' when calculating ``2.61 * 0.01'' while the correct answer is ``0.0261''.
We argue that this is primarily because the special operation \texttt{S-Mul} rarely occurs in compositional data, as we sample all values uniformly.
Whereas in the skill learning phase, we construct a large amount of training data for the \texttt{S-Mul} operation.
It indicates that the property of avoiding catastrophic forgetting mentioned in Section~\ref{sec:rq4} is conditional.
The distribution of atomic skills in compositional data being the same as in the prerequisite task.
The distribution of atomic skills within composite data needs to be consistent with the distribution in the prerequisite task, otherwise the phenomenon of catastrophic forgetting may still occur.

\subsection{Additional Examples on MWP}
\label{sec:ap_more}

We supplement \S~\ref{sec:rq1} with more complete examples, as shown in Tab.~\ref{tab:ap_mwp_example}.

\begin{table*}[t]
    \centering
    \small
    \begin{tabular}{ccccc}
    \toprule
       \# &  \multirow{1}{*}{Method} & Response & COT & CORRECT \\
    \midrule

\multirow{3}{*}{1} & \multirow{1}{*}{AL} & \multirow{1}{*}{\parbox{11cm}{\setlength{\fboxsep}{1pt}
495.3 + 996.8 = \colorbox{lightred}{1491.1}
}} & \multirow{1}{*}{-} & \multirow{1}{*}{\ding{55}} \\
\cdashlinelr{2-5}

 & \multirow{1}{*}{HCL} & \multirow{1}{*}{\parbox{11cm}{\setlength{\fboxsep}{1pt}
495.3 + 996.8 = 1492.1
}}  & \multirow{1}{*}{-} & \multirow{1}{*}{\ding{51}} \\
\midrule

\multirow{3}{*}{2} & \multirow{1}{*}{AL} & \multirow{1}{*}{\parbox{11cm}{\setlength{\fboxsep}{1pt}
6.8 \/ 0.1 = \colorbox{lightred}{68.33}
}} & \multirow{1}{*}{-} & \multirow{1}{*}{\ding{55}} \\
\cdashlinelr{2-5}

 & \multirow{1}{*}{HCL} & \multirow{1}{*}{\parbox{11cm}{\setlength{\fboxsep}{1pt}
68 \/ 0.1 = 68
}}  & \multirow{1}{*}{-} & \multirow{1}{*}{\ding{51}} \\
\midrule

\multirow{5}{*}{3} & \multirow{2}{*}{AL} & \multirow{2}{*}{\parbox{11cm}{\setlength{\fboxsep}{1pt}
18.4 * 9.945 = 18.4 * 9 + 18.4 * 0.9 + 18.4 * 0.04 + 18.4 * 0.005 = \colorbox{lightred}{166.6} + \colorbox{lightred}{16.66} + 0.736 + 0.092 = 183.26 + 0.736 + 0.092 = 183.996 + 0.092 = 184.088
}} & \multirow{2}{*}{\ding{51}} & \multirow{2}{*}{\ding{55}} \\ \\
\cdashlinelr{2-5} 

 & \multirow{2}{*}{HCL} & \multirow{2}{*}{\parbox{11cm}{\setlength{\fboxsep}{1pt}
18.4 * 9.945 = 18.4 * 9 + 18.4 * 0.9 + 18.4 * 0.04 + 18.4 * 0.005 = 165.6 + 16.56 + 0.736 + 0.092 = 182.16 + 0.736 + 0.092 = 182.896 + 0.092 = 182.988
}} & \multirow{2}{*}{\ding{51}} & \multirow{2}{*}{\ding{51}}  \\ \\
\midrule

\multirow{5}{*}{4} & \multirow{2}{*}{AL} & \multirow{2}{*}{\parbox{11cm}{\setlength{\fboxsep}{1pt}
69.4 * 81.8 = 69.4 * 80 + 69.4 * 1 + 69.4 * 0.8 = 5552 + \colorbox{lightred}{6.94} + 55.52 = 5618.94 + 55.52 = 5674.462
}} & \multirow{2}{*}{\ding{51}} & \multirow{2}{*}{\ding{55}} \\ \\
\cdashlinelr{2-5}

 & \multirow{2}{*}{HCL} & \multirow{2}{*}{\parbox{11cm}{\setlength{\fboxsep}{1pt}
69.4 * 81.8 = 69.4 * 80 + 69.4 * 1 + 69.4 * 0.8 = 5552 + 69.4 + 55.52 = 5621.4 + 55.52 = 5676.92
}}  & \multirow{2}{*}{\ding{51}} & \multirow{2}{*}{\ding{51}} \\ \\
\midrule

\multirow{5}{*}{5} & \multirow{2}{*}{AL} & \multirow{2}{*}{\parbox{11cm}{\setlength{\fboxsep}{1pt}
2.94 * 8.21 = 2.94 * 8 + 2.94 * 0.2 + 2.94 * 0.01 = \colorbox{lightred}{23.48} + 0.588 + 0.0294 = 24.068 + 0.0294 = 24.0974
}} & \multirow{2}{*}{\ding{51}} & \multirow{2}{*}{\ding{55}} \\ \\
\cdashlinelr{2-5}

 & \multirow{2}{*}{HCL} & \multirow{2}{*}{\parbox{11cm}{\setlength{\fboxsep}{1pt}
2.94 * 8.21 = 2.94 * 8 + 2.94 * 0.2 + 2.94 * 0.01 = 23.52 + 0.588 + 0.0294 = 24.108 + 0.0294 = 24.1374
}}  & \multirow{2}{*}{\ding{51}} & \multirow{2}{*}{\ding{51}} \\ \\
\midrule

\multirow{3}{*}{6} & \multirow{1}{*}{AL} & \multirow{1}{*}{\parbox{11cm}{\setlength{\fboxsep}{1pt}
539.25 - 418.4 = \colorbox{lightred}{120.8}
}} & \multirow{1}{*}{-} & \multirow{1}{*}{\ding{55}} \\
\cdashlinelr{2-5}

 & \multirow{1}{*}{HCL} & \multirow{1}{*}{\parbox{11cm}{\setlength{\fboxsep}{1pt}
539.25 - 418.4 = 120.85
}}  & \multirow{1}{*}{-} & \multirow{1}{*}{\ding{51}} \\

    \bottomrule
    \end{tabular}
    \caption{Example of the response in prerequisite task of different training strategies.  COT indicates whether the model answer with a COT format and CORRECT indicates whether the final answer is correct. We mark the areas that led to errors in red.}
    \label{tab:conv_example}
\end{table*}

\definecolor{lightred}{RGB}{255,200,200}
\definecolor{lightgreen}{RGB}{152, 251, 152}

\begin{table*}[t]
    \centering
    \small
    \begin{tabular}{cp{10.5cm}cc} 
    \toprule
        Method & \multicolumn{1}{c}{Response} & SKILL & CORRECT \\
    \midrule
     \multirow{3}{*}{{Question}} & Jill gets paid \$631 per hour to teach and \$30 to be a cheerleading coach. If she works 38 weeks a year, 32 hours a week as a teacher and 15 hours a week as a coach, what's her annual salary? \\
    \cdashlinelr{1-4}
      \multirow{5}{*}{ST} &  Jill works 38 weeks a year * 32 hours a week = 1304 hours a year as a teacher. She earns 1304 hours a year * \$631 a hour = \colorbox{lightred}{\$800800} a year as a teacher. As a coach, Jill works 15 hours a week * 38 weeks a year = 570 hours a year. She earns 570 hours a year * \$30 a hour = \$17100 a year as a coach. Therefore, Jill's annual salary is \$17100 + \$800800 = \$802900.
     & \multirow{5}{*}{\ding{55}} & \multirow{5}{*}{\ding{55}} \\
     \cdashlinelr{1-4}
      \multirow{5}{*}{AL} & As a teacher, Jill works 38*32  = 38 * 30 + 38 * 2 = 1140 + 76 = 1216 weeks. She earns 1216*\$631 = 1216 * 600 + 1216 * 30 + 1216 * 1 = \colorbox{lightred}{731600} + 36480 + 1216 = 768080 + 1216 = 769296. As a coach, Jill works 38*15 = 38 * 10 + 38 * 5 = 380 + 190 = 570 weeks. She earns 570*\$30 = 570 * 30 + 570 * 0 = 17100 + 0 = 17100. Annually, she earns \$769296+\$17100 = \$786396
     & \multirow{5}{*}{\ding{55}} & \multirow{5}{*}{\ding{55}} \\
     \cdashlinelr{1-4}
      \multirow{7}{*}{HCL} & Jill earns 38 weeks * 32 hours/week = 38 * 30 + 38 * 2 = 1140 + 76 = 1216 hours a year as a teacher. She earns 1216 hours/year * \$631/hour = 1216 * 600 + 1216 * 30 + 1216 * 1 = 729600 + 36480 + 1216 = 766080 + 1216 = 767296 a year as a teacher. She earns 15 hours/week * \$30/hour = 15 * 30 = 15 * 30 + 15 * 0 = 450 + 0 = 450 a week as a coach. In a year, she earns \$450/week * 38 weeks/year = 450 * 38 = 450 * 30 + 450 * 8 = 13500 + 3600 = 17100 as a coach. Her total salary is \$767296 + \$17100 = 784396. 
      & \multirow{7}{*}{\ding{51}} & \multirow{7}{*}{\ding{51}} \\

    \midrule
    
     \multirow{2}{*}{{Question}} & A raspberry bush has 911 clusters of 637 fruit each and 797 individual fruit scattered across the bush. How many raspberries are there total? \\
    \cdashlinelr{1-4}
      \multirow{3}{*}{ST} & There are 911 clusters of raspberries, each with 637 fruit, so there are 911 * 637 = \colorbox{lightred}{545702} raspberries in the clusters. There are also 797 individual raspberries, so there are a total of 545702 + 797 = 553699 raspberries on the bush. 
      & \multirow{3}{*}{\ding{55}} & \multirow{3}{*}{\ding{55}} \\
     \cdashlinelr{1-4}
     \multirow{4}{*}{AL} & There are 911 clusters of raspberries on the bush, so there are 911 * 637 = 911 * 600 + 911 * 30 + 911 * 7 = 546600 + 27330 + 6377 = 573930 + 6377 = 580307 raspberries in the clusters. There are also 797 individual raspberries scattered across the bush, so there are a total of 580307 + 797 = 581104 raspberries.
     & \multirow{4}{*}{\ding{51}} & \multirow{4}{*}{\ding{51}} \\
     \cdashlinelr{1-4}
     \multirow{4}{*}{HCL} & First find the total number of fruit in the clusters: 911 clusters * 637 fruit/cluster = 911 * 600 + 911 * 30 + 911 * 7 = 546600 + 27330 + 6377 = 573930 + 6377 = 580307 fruit. Then add the number of individual fruit to find the total number of fruit: 580307 fruit + 797 fruit = 581104 fruit. 
     & \multirow{4}{*}{\ding{51}} & \multirow{4}{*}{\ding{51}} \\

    \midrule
    
    \multirow{2}{*}{{Question}} &  A glass of milk is 668 ounces of milk.  John drinks 671 glasses of milk.  If milk has 949 calories per ounce how many calories did he consume? \\
    \cdashlinelr{1-4}
     \multirow{2}{*}{ST} & He drank 671*668=\colorbox{lightred}{405508} ounces of milk. That means he consumed 405508*949=36545720 calories. 
     & \multirow{2}{*}{\ding{55}} & \multirow{2}{*}{\ding{55}} \\
     \cdashlinelr{1-4}
 \multirow{4}{*}{AL} & He drank 668*671= 668 * 600 + 668 * 70 + 668 * 1 = 400800 + 46760 + 668 = 447560 + 668 = 448228 ounces of milk. That means he consumed 448228*949 = 448228 * 900 + 448228 * 40 + 448228 * 9 = \colorbox{lightred}{393405200} + 17929120 + \colorbox{lightred}{3934052} = 411334320 + 3934052 = 415268372 calories.
     & \multirow{4}{*}{\ding{55}} & \multirow{4}{*}{\ding{55}} \\
     \cdashlinelr{1-4}
     \multirow{4}{*}{HCL} & He drank 671*668 = 671 * 600 + 671 * 60 + 671 * 8 = 402600 + 40260 + 5368 = 442860 + 5368 = 448228 ounces of milk. So he consumed 448228*949= 448228 * 900 + 448228 * 40 + 448228 * 9 = 403405200 + 17929120 + 4034052 = 421334320 + 4034052 = 425368372 calories.  & \multirow{4}{*}{\ding{51}} & \multirow{4}{*}{\ding{51}} \\

    \bottomrule
    \end{tabular}
    \caption{Example of the response in complex reasoning task of different training strategies.  SKILL indicates whether the model fail in atomic skills (arithmetic). CORRECT indicates whether the final answer is correct. We mark the areas that led to errors in red.}
    \label{tab:ap_mwp_example}
\end{table*}

\end{document}